\documentclass[runningheads]{llncs}
\usepackage{eccv}
\usepackage{eccvabbrv}
\usepackage{graphicx}
\usepackage{booktabs}
\usepackage[accsupp]{axessibility}
\usepackage{amsmath}
\usepackage{wrapfig}
\usepackage{multirow}
\usepackage{colortbl}
\usepackage{algorithm} 
\usepackage[ruled,vlined,algo2e]{algorithm2e}
\newcommand{\myparagraph}[1]{\vspace{1mm}\noindent{\bf #1}}

\definecolor{first}{rgb}{1.0, 0.6, 0.6}    
\definecolor{third}{rgb}{1.0, 1.0, 0.6} 
\definecolor{second}{rgb}{1.0, 0.8, 0.6} 
\usepackage{hyperref}
\usepackage{orcidlink}

\begin{document}

\title{Delaunay Canopy: Building Wireframe Reconstruction from Airborne LiDAR Point Clouds via Delaunay Graph} 
\titlerunning{Delaunay Canopy}

\author{
Donghyun Kim\inst{1, 2} \and
Chanyoung Kim\inst{1, 2}\and
Youngjoong Kwon\inst{1} \and
Seong Jae Hwang\inst{2}
}
\authorrunning{D. Kim et al.}

\institute{
Emory University, Atlanta GA, USA \\
\email{\{donghyun.kim, chanyoung.kim, youngjoong.kwon\}@emory.edu}
\and
Yonsei University, Seoul, Republic of Korea \\
\email{seongjae@yonsei.ac.kr}
}

\maketitle

\begin{center}
    \centering
    \includegraphics[width=\textwidth]{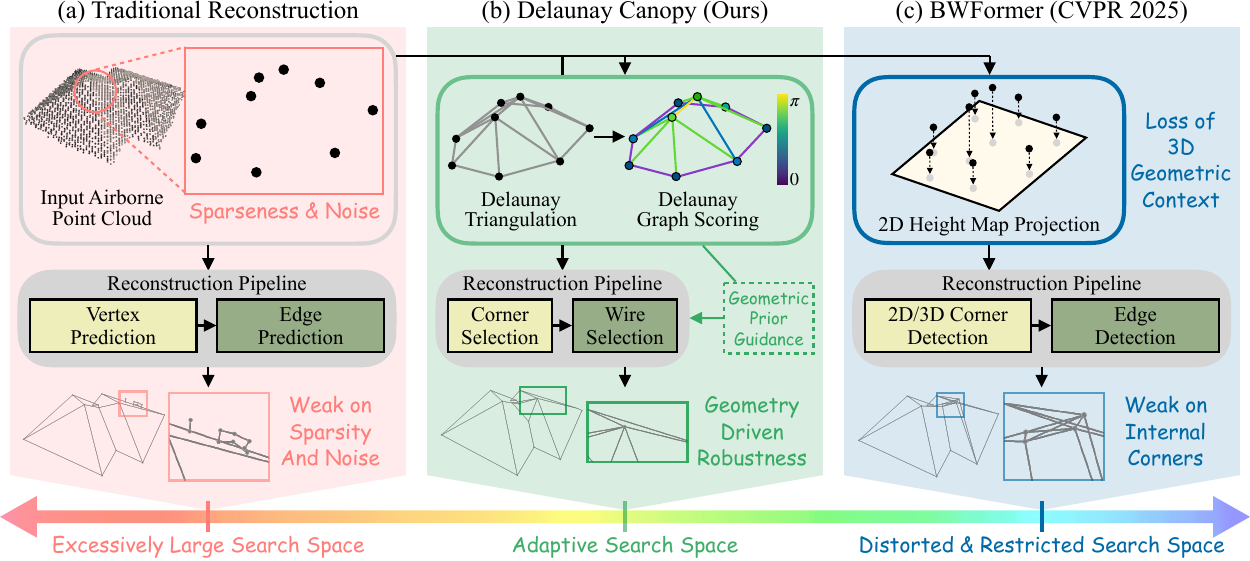}
    \captionof{figure}{We present Delaunay Canopy for building wireframe reconstruction. Compared to traditional methods with overly large search space (left) and 2D height map projection that loses 3D context (right), our approach scores a Delaunay graph with geometric priors to select corners and wires, yielding an adaptive (\ie, geometrically context‑aware) search space that enables precise wireframe reconstruction.}
    \label{fig:teaser}
\end{center}

\begin{abstract}
    Reconstructing building wireframe from airborne LiDAR point clouds yields a compact, topology-centric representation that enables structural understanding beyond dense meshes.
    Yet a key limitation persists: conventional methods have failed to achieve accurate wireframe reconstruction in regions afflicted by significant noise, sparsity, or internal corners. 
    This failure stems from the inability to establish an adaptive search space to effectively leverage the rich 3D geometry of large, sparse building point clouds.
    In this work, we address this challenge with Delaunay Canopy, which utilizes the Delaunay graph as a geometric prior to define a geometrically adaptive search space. 
    Central to our approach is Delaunay Graph Scoring, which not only reconstructs the underlying geometric manifold but also yields region-wise curvature signatures to robustly guide the reconstruction.
    Built on this foundation, our corner and wire selection modules leverage the Delaunay-induced prior to focus on highly probable elements, thereby shaping the search space and enabling accurate prediction even in previously intractable regions.
    Extensive experiments on the Building3D Tallinn city and entry-level datasets demonstrate state-of-the-art wireframe reconstruction, delivering accurate predictions across diverse and complex building geometries.
\end{abstract}
\section{Introduction}
\label{sec:introduction}

3D building data serves as a cornerstone across diverse domains such as the metaverse, XR (Extended Reality)~\cite{ar_reference1, ar_reference2, ar_reference3}, autonomous driving~\cite{autonomous_driving_reference1, wang2016torontocity,huang2018apolloscape}, gaming~\cite{gaming_ref1, gaming_ref2}, and robotics~\cite{robotics_ref1, ramakrishnan2021hm3d, dehghan2021arkitscenes}.
Acquiring real-world 3D building data typically begins by scanning buildings with LiDAR sensors to obtain a 3D point cloud.
Although point clouds are highly accurate representations~\cite{pointconv, dgcnn, curvenet, point_transformer_v2, point_transformer_v3}, their inherent sparsity and irregularity, coupled with data size that scales with object size, may pose challenges for their use as lightweight representations for the aforementioned applications~\cite{mao2019interpolated}.
Consequently, there is a need for techniques that reconstruct 3D building point clouds into precise and lightweight representations, and the \textit{wireframe} has increasingly become the format that best satisfies these requirements.
Nevertheless, robustly dealing with the point cloud’s sparsity and irregularity remains a key challenge, as wireframe reconstruction hinges on accurately recovering sharp edges, corners, and topological connectivity, which are highly sensitive to noise and missing data.

Despite the glaring challenge, studies on reconstructing 3D building point clouds into wireframes have advanced rapidly. 
Beginning with the first network-based Point2Roof~\cite{point2roof}, subsequent studies~\cite{building3d, pc2wf} in wireframe reconstruction have adopted a pipeline that first predicts corners of the wireframe and then predicts wires. 
This pipeline simplifies the overall problem by decomposing a complex task into smaller modules; however, it fails to account for the enormous scale, sparsity, and noise of building point clouds (\cref{fig:teaser}(a)), resulting in an \textit{excessively large search space} that challenges accurate reconstruction.

\begin{wrapfigure}{r}{0.55\textwidth}
    \centering
    \vspace{-15pt} 
    \includegraphics[width=\linewidth]{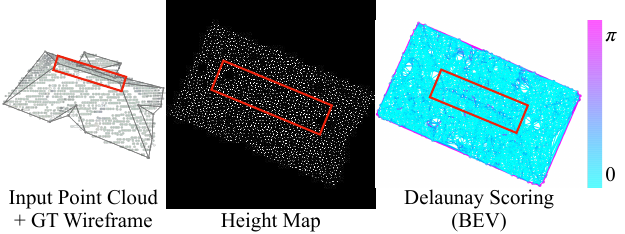}
    \vspace{-15pt} 
    \captionof{figure}{\textbf{Comparison of geometric context between 2D heightmap and Delaunay scoring.} While the 2D heightmap collapses 3D structural details and obscures internal corners (red box), our Delaunay scoring effectively preserves the underlying geometric context.}
    \label{fig:heightmap}
    \vspace{-15pt} 
\end{wrapfigure}
To address this, BWFormer~\cite{bwformer} introduced a 2D-to-3D corner detection approach in which the point cloud is projected onto a 2D height map (\cref{fig:teaser}(c)). 
In the 2D dimension, corners are detected first, and these detections are then used to infer the 3D corners.
While this dimensionality reduction constrains the search space, such constraint becomes \textit{overly restrictive}, leading to critical limitations due to the loss of intricate 3D contextual information.
Most critically, BWFormer struggles to detect interior corners, particularly because it restricts the search space to a 2D BEV (Bird's Eye View) perspective, thereby failing to fully leverage given 3D information (\cref{fig:heightmap}).
A related drawback is that projecting the continuous 3D point cloud into a 2D grid representation distorts the inherent 3D search space, as the 3D coordinates must be quantized into integer indices.
Thus, in 3D building wireframe reconstruction, substantial effort must be paid to establish an \textit{adaptive search space} over point clouds that is informed by the underlying 3D context, ensuring compactness to suppress the detrimental effects of sparsity and noise while preserving adequate scope to encompass all geometrically valid structures.

\begin{figure}[t]
    \centering
    \includegraphics[width=0.8\linewidth]{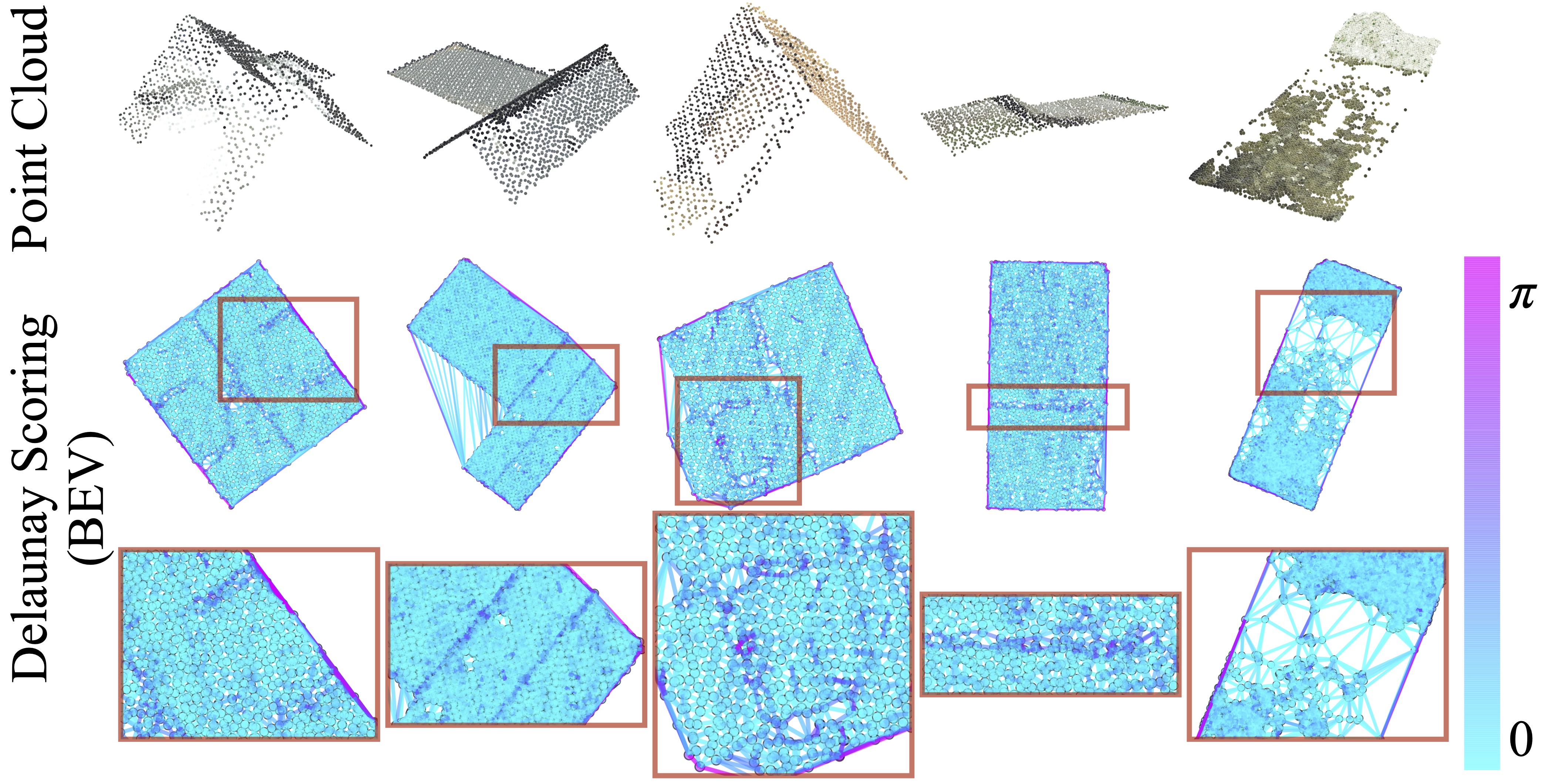}
    \caption{\textbf{Visualization of the edge-wise dihedral angle and vertex-wise corner score}, derived through Delaunay graph scoring across input point clouds, projected onto the Bird's-Eye View (BEV).
    The results demonstrate a robust capability to accurately capture the underlying surface and the curvature of the corresponding region. Notably, the scoring adeptly represents even the most minute degrees of surface curvature (fourth column) and maintains efficacy even when the point cloud is highly sparse (fifth column).}
    \label{fig:delaunay_scoring}
\end{figure}

Our 3D building wireframe reconstruction framework, \textit{Delaunay Canopy}, aims to ensure an adaptive search space by leveraging the geometric prior of the Delaunay graph (\cref{fig:teaser}(b)).
The Delaunay graph, constructed via Delaunay triangulation of the input point cloud, captures 3D geometric information by approximating the underlying surface of a building roof in 3D space, irrespective of sparsity and scale.
Motivated by the fact that critical points (\eg, roof corners) exhibit high local curvature, we propose \textit{Delaunay graph scoring}, the core approach of our framework designed to extract curvature signatures from the graph for guiding reconstruction (\cref{fig:delaunay_scoring}).
Specifically, by computing the \textit{dihedral angle} along an edge (shared by two triangular faces), one can infer the local curvature of the underlying surface at that location. 
These dihedral angles are then aggregated for each vertex to calculate a \textit{corner score}, representing the degree of local curvature associated with that vertex.

We leverage the curvature signatures from Delaunay graph scoring to predict wireframe corners and wires.
The resulting Delaunay-induced prior concentrates the corner and wire selection on highly probable elements (\ie, candidates in high-curvature regions), scoping the search space and enabling accurate prediction in challenging regions (\eg, internal corners).
With this prior in place, we adaptively narrow the search space by first sampling and classifying input points with high corner scores as corner candidates, and then train a corner selection module to predict the final corners.
For wire selection, we generate wire candidates from predicted corners and assign a \textit{path score} by summing dihedral angles along the shortest Delaunay graph path between a candidate's endpoints.
Subsequently, we scale the query corresponding to each candidate based on the path score, and the wire selection module is trained to amplify information about promising candidates, thereby selecting valid wires.
Through a carefully designed pipeline, our Delaunay Canopy adaptively calibrates the search space using Delaunay graph scoring, a strong geometric cue, which in turn yields accurate final wireframe through a corner and wire selection module.

The main contributions of this work are:
\begin{itemize}
    \item We propose Delaunay graph scoring, a method that applies Delaunay triangulation to the point cloud to quantify local curvature, transforming the underlying surface geometry into essential curvature signatures.
    
    \item We introduce a corner and wire selection pipeline that leverages a geometric cue to adaptively adjust the search space, enabling the model to focus on probable candidates and achieve precise results in challenging regions.

    \item We demonstrate the effectiveness of Delaunay Canopy on the Building3D dataset and achieve state-of-the-art results, supported by a comprehensive experimental study.
\end{itemize}
\section{Related Work}
\label{sec:related_work}

\myparagraph{Building Reconstruction.}
Building reconstruction plays a crucial role in vision applications~\cite{bwformer,point2roof,li20243d,Mahmud_2020_CVPR,pbwr,point2building,buildingnet}. Recently, some methods such as City3D~\cite{city3d} and Kinetic Shape Reconstruction~\cite{kinetic} have been proposed to directly output surface meshes by enforcing global plane primitives and topological correctness. 
While these approaches guarantee watertightness, they often over-close sparse topology by forcing global consistency.
In contrast, wireframe representations offer a more flexible alternative by prioritizing structural correctness and superior editability over forced watertightness, focusing on concise, topology-centric descriptions~\cite{point2roof,bwformer,pbwr}.
Pioneering work~\cite{point2roof} detects corners and predicts wires to assemble wireframes, but it often misses corners due to the sparsity of point clouds. 
PBWR~\cite{pbwr} outputs edges in a direct fashion, yet it still requires post-processing and falls short of being a unified pipeline.
More recently, BWFormer~\cite{bwformer} introduced a 2D-to-3D corner detection approach that projects the point cloud to a height map, detects 2D corners, and then infers 3D corners. However, this projection quantizes 3D coordinates on a 2D grid, introduces geometric errors, requires corner detection in both 2D and 3D with doubled computational cost, and overly constrains the search space that it underuses 3D cues and struggles with interior corners in BEV.
In contrast, our model operates explicitly in 3D with a Delaunay graph representation, while leveraging geometric cues for adaptive search space guidance.

\myparagraph{Delaunay Graph for 3D Reconstruction.}
Delaunay triangulation is a classical geometric tool for organizing unstructured 3D point clouds into surface and has been widely used in reconstruction tasks~\cite{learning_delaunay,son2024dmesh,deepdt}. 
Its graph structure provides a powerful geometric prior, as edges naturally encode local neighborhood relationships and capture nearest-neighbor connections~\cite{zhang2023dmnet}. 
This property confines the search space to spatially meaningful candidate edges on the underlying surface. 
Classical methods like the Ball-Pivoting Algorithm~\cite{bernardini2002ball} have successfully leveraged this connectivity to guide surface growth and ensure manifold consistency.
In our approach, we exploit the Delaunay graph to define the search space for wireframe reconstruction. 
Each triangulation edge represents a potential roof edge, and its dihedral angle is used to estimate local curvature. 
Aggregating these angles yields a corner score, enabling the selection of high-curvature corner candidates. 
This Delaunay-based formulation ensures a geometrically informed and computationally efficient search for structural elements grounded in the input geometry.
\section{Method}
\label{sec:method}

In this section, we introduce our 3D building reconstruction framework, \textit{Delaunay Canopy}. 
We first describe the \textit{Delaunay graph scoring} (\cref{subsec:delaunay_graph}), which serves as a foundational module providing a 3D geometric guidance within our framework.
Subsequently, we present the corner and wire selection pipeline (\cref{subsec:corner_selection_and_wire_selection}), guided by the geometric prior derived from Delaunay graph scoring.

\subsection{Delaunay Graph Scoring}
\label{subsec:delaunay_graph}
Key features in 3D building (\eg, corners) exhibit pronounced local curvature.
This observation raises the question: \emph{Can a curvature metric be leveraged as a strong prior to inform the topology and guide wireframe reconstruction?}
To investigate, we first introduce \textit{Delaunay graph scoring}, which approximates the underlying surface of the point cloud and computes region-wise curvature metrics.

\begin{figure*}[t]
    \centering
    \includegraphics[width=\linewidth]{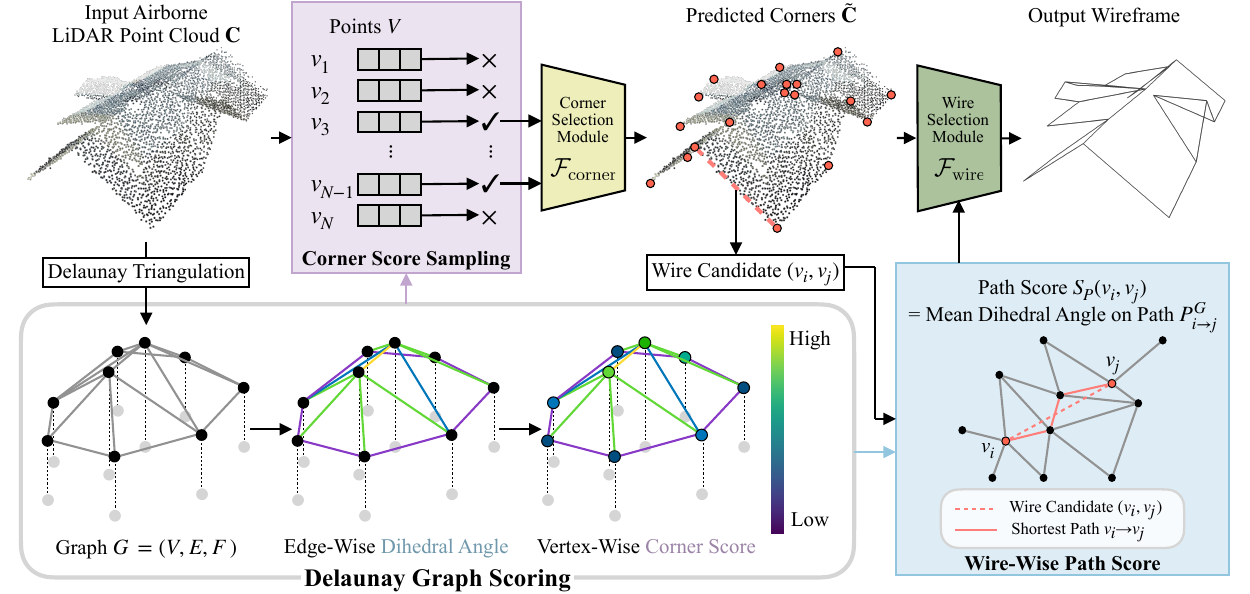}
    \caption{\textbf{Overall pipeline of Delaunay Canopy.} 
    The pipeline begins by utilizing Delaunay triangulation to construct the point cloud into a Delaunay graph. The method then sequentially computes metrics that represent the local curvature around each element, proceeding in the order of face, edge, and vertex. This process encapsulates rich 3D information and introduces a judicious search space into the pipeline.
    }
    \label{fig:Overall_pipeline}
\end{figure*}

Unlike many conventional 3D object point clouds~\cite{semantic3d,3dshapenets}, airborne LiDAR scans of building roofs inherently represent surfaces with open topologies: for a given $(x, y)$ coordinate, the $z$ value is mapped without occlusion. 
This property enables us to estimate an implicit representation of the point cloud's underlying surface.
Building upon this, we apply \textit{Delaunay Triangulation} to the airborne LiDAR building point cloud to approximate its underlying surface. 
This process generates a Delaunay graph $G = (V, E, F)$, which holds both mesh and graph properties and comprises vertices $V$, edges $E$, and faces $F$.
To estimate the curvature of a specific region, we derive geometric priors specifically for curvature characterization upon the Delaunay graph $G$.
These priors are assigned to the graph’s components as follows: (\romannumeral1) faces $F$ via \textit{Face Normal Computation}, (\romannumeral2) edges $E$ via \textit{Edge-Wise Dihedral Angle}, and (\romannumeral3) vertices $V$ via \textit{Vertex-Wise Corner Score} (see \cref{fig:Overall_pipeline}). 
We describe each step in a detail below.

\myparagraph{(\uppercase\expandafter{\romannumeral1}) Face Normal Computation.}
\label{subsubsec:face_normal_computation}
For every face $f = (v_a, v_b, v_c)$ within a Delaunay graph $G$, let the 3D coordinates of its three constituent vertices be $\textbf{v}_a$, $\textbf{v}_b$, and $\textbf{v}_c$. The normal vector $\mathbf{n}_f$ for each face $f$ is computed as $\textbf{n}_f = (\textbf{v}_b - \textbf{v}_a) \times (\textbf{v}_c - \textbf{v}_a) \in \mathbb{R}^3$.
To ensure consistent normal orientation, if the calculated normal vector points downward (\ie, the $z$-component of $\mathbf{n}_f$ is less than zero), its direction is inverted to $-\mathbf{n}_f$.

\myparagraph{(\uppercase\expandafter{\romannumeral2}) Edge-Wise Dihedral Angle.}
\label{subsubsec:edge_wise_dihedral_angle}
With the face normals in hand, we now compute curvature-related geometric information at the edge level by computing the \textit{dihedral angle} for each edge. 
For every edge $e$, we define the set of two adjacent faces as follows: $F(e) = \{f \in F \vert e \subset f \}$.
Given an edge $e$ with its adjacent faces $F(e) = \{f^1, f^2\}$, and their respective normal vectors $\mathbf{n}_1$ and $\mathbf{n}_2$, the dihedral angle $\theta_e$ is computed as
\begin{equation}
    \small
    \theta_e = \arccos \left( \frac{\mathbf{n}_1 \cdot \mathbf{n}_2}{\Vert\mathbf{n}_1\Vert \ \Vert \mathbf{n}_2 \Vert} \right) \in \left[ 0, \pi \right].
\end{equation}
A large dihedral angle at an edge signifies a sharp fold or corner where the two adjacent faces meet, whereas a small angle indicates that the region is relatively flat. For edges situated on the periphery and thus abutting only a single surface, the dihedral angle $\theta_e$ is configured to $\pi$.

\myparagraph{(\uppercase\expandafter{\romannumeral3}) Vertex-Wise Corner Score.}
\label{subsubsec:vertex_wise_corner_score}
At the most granular level of the graph, the vertex, we compute a \textit{corner score} using curvature information from its incident edges. 
For a vertex $v$, its corner score $S_c(v)$ is determined by aggregating the dihedral angles of all incident edges $E(v) =\{e \in E \vert v \in e\}$, defined as
\begin{equation}
    \small
    S_c(v) = \frac{1}{\vert E(v) \vert} \sum_{e \in E(v)}\theta_e.
\end{equation}
A high corner score implies the presence of numerous edges with high dihedral angles, suggesting that the local region around that vertex has significant curvature and the vertex is highly likely to be an actual corner of the building.

\begin{figure}[t]
    \centering
    \includegraphics[width=0.8\linewidth]{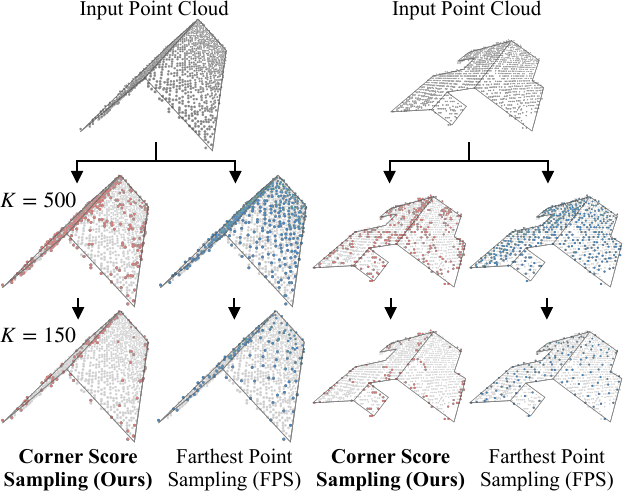}
    \caption{
    \textbf{Comparison between Corner Score Sampling (ours) and Farthest Point Sampling (FPS).} Resultant point sets after sequentially sampling 500 points ($K$ = 500) and then 150 points ($K$ = 150) with each method are shown.
    For enhanced clarity, the ground truth wireframe is visualized. While FPS performs a uniform sampling, our corner score sampling method retains points located in areas of pronounced curvature (\eg, roof corners and edges), making them highly probable corner candidates.
    }
    \label{fig:corner_score_sampling}
\end{figure}

\subsection{Corner and Wire Selection}
\label{subsec:corner_selection_and_wire_selection}
Delaunay graph scoring, described in Sec.~\ref{subsec:delaunay_graph}, captures local 3D geometric information and provides region-wise curvature metrics.
These metrics enable adaptive scoping  of the search space for predicting the final corners and wires of the wireframe.
Let us describe how the wireframe is reconstructed (see \cref{fig:Overall_pipeline}) by utilizing Delaunay-informed priors through the following stages: corner selection (\cref{subsubsec:corner_selection}) and wire selection (\cref{subsubsec:wire_selection}).

\subsubsection{Corner Selection.}
\label{subsubsec:corner_selection}
Corners define the key geometric structure of a building, marking where planes meet or change direction. However, airborne point clouds contain dense planar samples, most of which are not structually meaningful. The goal of the corner selection stage is therefore to (\romannumeral1) isolate a small, geometry-aware subset of points that are likely to represent true corners, and then (\romannumeral2) refine them into the final set of corners.

\myparagraph{(\uppercase\expandafter{\romannumeral1}) Corner Score Sampling.}
We first narrow the search space by sampling vertices that exhibit high corner scores, computed using the \textit{vertex-wise corner score} (see \cref{subsubsec:vertex_wise_corner_score}).
Specifically, from the input point cloud $\mathbf{C} \in \mathbb{R}^{N \times 3},$
we select the top $K$ points with the highest corner scores, forming the subset $\mathbf{C}' \in \mathbb{R}^{K \times 3}.$
As illustrated in \cref{fig:corner_score_sampling}, decreasing $K$ yields sampled points that are progressively more densely located along the roof's boundaries and junctions, indicating a higher likelihood of representing actual building corners.
This step effectively removes redundant planar points, and yields an \textit{adaptive and focused search space} guided by Delaunay priors.

\myparagraph{(\uppercase\expandafter{\romannumeral2}) Corner Selection Module.}
Then, the subset \(\mathbf{C}'\) is subsequently passed into the corner selection module, expressed as a learnable function \(\mathcal{F}_{\text{corner}}(\cdot)\), which is trained to refine \(\mathbf{C}'\) to the final set of predicted corners \(\tilde{\mathbf{C}} \in \mathbb{R}^{M \times 3}\), formally as $\tilde{\mathbf{C}} = \mathcal{F}_{\text{corner}}(\mathbf{C}')$.
Details of the training process are provided in the supplementary.

\subsubsection{Wire Selection.}
\label{subsubsec:wire_selection}
Once the corners are identified, the next step is to determine which pair of corners form valid wires using a differentiable (\romannumeral1) \textit{wire selection module}.
However, treating wire candidates that possess distinct geometric properties in the same manner can potentially lead to distorted information processing.
Thus, we introduce (\romannumeral2) \textit{wire-wise path score}, which measures geometric coherence, and use it within (\romannumeral3) \textit{prior-based query scaling} to dynamically weight each wire query.
Let us first describe the wire selection module.

\myparagraph{(\uppercase\expandafter{\romannumeral1}) Wire Selection Module.}
The wire selection module $\mathcal{F}_{\text{wire}}$ takes all $\binom{M}{2}$ wire candidates $\mathcal{W} = \{ w_{ij} = (v_i, v_j) \mid 1 \le i < j \le M \}$ as input, which are constructed from every possible pair of the $M$ predicted corners, and determines whether each candidate $w_{ij}$ constitutes a valid wire. 
The module is built upon a multi-layer Transformer decoder~\cite{detr} that classifies each wire candidate using its spatial configuration.
Each candidate $w_{ij}$ is represented as a query and, for layer $l$, is encoded as a wire-specific query vector $Q_l^w$, which is initialized from the 3D coordinates of its endpoints $v_i$ and $v_j$ at the first decoder layer.
Within each decoder layer, the model performs self-attention among all queries and edge attention~\cite{bwformer}, where features are uniformly sampled along the edge and aggregated via max-pooling. After all layers have been processed, the final query embeddings are passed through an MLP to produce a validity score for each wire. 
Additional details are included in the supplementary.
Although this effectively models relationships among wire candidates, it treats all queries as equally important. In reality, wire candidates vary substantially in geometric plausibility (\eg, some follow clear structural ridges, while others cross unstructured regions). Giving all queries equal weight may therefore divert model capacity toward implausible wires. To address this gap, we incorporate the Delaunay-derived \textit{path score} as a dynamic weighting factor.

\myparagraph{(\uppercase\expandafter{\romannumeral2}) Wire-Wise Path Score.}
For a wire candidate $w_{ij}$ connecting vertices $v_i$ and $v_j$, we first find the shortest path $P^G_{i \rightarrow j}$ between these two vertices on the Delaunay graph $G$. The edges along this path, denoted as $L(P^G_{i \rightarrow j})$, share the same local region as the wire candidate. 
By averaging the dihedral angles of these edges, we can compute the path score $S_P(v_i, v_j)$, which represents the wire curvature:
\begin{equation}
    \small
    S_P(v_i, v_j) = \frac{1}{\vert L(P^G_{i \rightarrow j}) \vert} \sum_{e \in L(P^G_{i \rightarrow j})}\theta_e.
\end{equation}
The path score subsequently serves as the primary value used by the wire selection module to determine valid wires.

\myparagraph{(\uppercase\expandafter{\romannumeral3}) Prior-based Query Scaling.}
At each decoder layer $l$, the query vector $Q_l^w$ of a wire candidate $w$ is modulated by its path score $S_P^w$. The score is mapped through a sigmoid activation $s^w = \sigma(S_P^w)$ to produce a scale factor that is applied element-wise: ${Q'}_l^w = Q_l^w \odot s^w$.
This modulation rescales query magnitudes, damping low-score wires and emphasizing high-confidence ones. Consequently, the model's attention is focused on geometrically plausible structures, leading to stable and accurate wire predictions.
\section{Experiments}
\label{sec:experiments}

In this section, we first outline the experimental setup, including the datasets, evaluation metrics, and implementation details. We then present both quantitative and qualitative results, followed by analyses, including ablation studies and a detailed examination of each proposed component. 

\subsection{Experimental Setup}
\myparagraph{Datasets.}
To evaluate the wireframe reconstruction performance of the Delaunay Canopy, we utilize the Building3D dataset~\cite{building3d}, which encompasses both the Tallinn city and entry-level datasets. 
Crucially, the complete complement of input point cloud and ground truth (GT) wireframe pairs for both datasets is not available in an entirely open-source format; only a subset has been publicly disclosed.
Consequently, from the Tallinn city dataset, we employ the 32,618 publicly available input-GT pairs, which are randomly partitioned into a training set of 30,000 and a test set of 2,618. For the entry-level dataset, we employ the total of 5,698 disclosed pairs, splitting them randomly into 5,000 for the training set and the remaining 698 for the test set.
To ensure a fair comparison with established benchmarks, we also evaluate our model following the Building3D official leaderboard protocols as detailed in the supplementary material.

\myparagraph{Metrics.}
We evaluate our method using eight common metrics for building wireframe reconstruction. 
Distance metrics include Wireframe Edit Distance (WED) and Average Corner Offset (ACO). 
We also incorporate corner-centric metrics (Corner Precision (CP), Corner Recall (CR), and Corner F1 (CF1)) and wire-level metrics (Edge Precision (EP), Edge Recall (ER), and Edge F1 (EF1)).
In the tables, the corner and wire metrics are presented as percentages (\%).

\myparagraph{Baselines.}
The representative baselines selected for comparison are PC2WF~\cite{pc2wf}, Point2Roof~\cite{point2roof}, and BWFormer~\cite{bwformer}. 
Additionally, a baseline setting utilizing PointNet~\cite{pointnet}, as introduced in Building3D~\cite{building3d}, is also implemented and evaluated.
Consistent with prior works~\cite{pbwr, bwformer}, evaluation of \cite{pc2wf} utilizes a model pretrained on ABC dataset~\cite{abc}.
Notably, to resolve potential performance discrepancies, we train and evaluate Point2Roof~\cite{point2roof} on the identical datasets used for all other baselines, whereas prior works~\cite{building3d, pbwr, bwformer} often evaluated it using models pretrained on synthetic data.
Other prior works, such as PBWR~\cite{pbwr} and various other methods (\eg, linear self-supervised) reported in Building3D~\cite{building3d}, can not be utilized for performance assessment due to the unavailability of their code. 
To establish a more diverse set of baselines, we measure performance by substituting diverse point cloud processing backbones (PointTransformer~\cite{point_transformer}, PointMLP~\cite{pointmlp}, PointNeXt~\cite{pointnext}, and PointMeta~\cite{pointmeta}) for the feature extraction component within the Point2Roof~\cite{point2roof} pipeline, analogous to the methodology employed in Building3D~\cite{building3d}.

\myparagraph{Implementation Details.}
All airborne LiDAR point clouds utilized as input are scaled to a range between 0 and 256.
Our model is trained for 300 epochs using AdamW with an initial learning rate of 0.0001. 
Within our framework, the corner selection module and the wire selection module both use a Transformer-based architecture with 6 layers each.
In the corner selection process, the number of points sampled via \textit{corner score sampling} is set to 150.
Additionally, within the wire selection module, the \textit{prior-based query scaling} is only implemented across the first three transformer layers.

\renewcommand{\arraystretch}{1.1}
\begin{table*}[t]
    \caption{\textbf{Quantitative comparisons on the Building3D Tallinn city and entry-level datasets.} The notation $\texttt{MODEL}^*$ signifies a model acting as the feature extractor within the Point2Roof~\cite{point2roof} pipeline. \colorbox{first}{Best}, \colorbox{second}{second best}, and \colorbox{third}{third best} are highlighted.}
    \centering
    \setlength{\aboverulesep}{0pt}
    \setlength{\belowrulesep}{0pt}
    \setlength{\tabcolsep}{1.1pt}
    \resizebox{\textwidth}{!}{
    \begin{tabular}{l cc ccc ccc cc ccc ccc}
        \toprule
        \multirow{3}{*}{Method} & \multicolumn{8}{c}{Tallinn City} & \multicolumn{8}{c}{Entry-Level} \\
        \cmidrule(lr){2-9} \cmidrule(lr){10-17}
        & \multicolumn{2}{c}{Distance} & \multicolumn{3}{c}{Corner} & \multicolumn{3}{c}{Wire} & \multicolumn{2}{c}{Distance} & \multicolumn{3}{c}{Corner} & \multicolumn{3}{c}{Wire} \\
        \cmidrule(lr){2-3} \cmidrule(lr){4-6} \cmidrule(lr){7-9} \cmidrule(lr){10-11} \cmidrule(lr){12-14} \cmidrule(lr){15-17}
        & WED $\downarrow$ & ACO $\downarrow$ & CP $\uparrow$ & CR $\uparrow$ & CF1 $\uparrow$ & EP $\uparrow$ & ER $\uparrow$ & EF1 $\uparrow$ & WED $\downarrow$ & ACO $\downarrow$ & CP $\uparrow$ & CR $\uparrow$ & CF1 $\uparrow$ & EP $\uparrow$ & ER $\uparrow$ & EF1 $\uparrow$ \\
        \midrule
        Building3D (PointNet)~\cite{pointnet} & 0.264 & 0.241 & 88.2 & 77.6 & 82.6 & 80.4 & 71.9 & 75.9 & 0.262 & 0.239 & 89.1 & 77.8 & 83.1 & 80.7 & 72.7 & 76.5 \\ 
        PC2WF~\cite{pc2wf} & 0.554 & 0.508 & 21.4 & 50.5 & 30.1 & 2.8 & 16.7 & 4.8 & 0.490 & 0.422 & 28.5 & 16.3 & 20.7 & 3.1 & 19.1 & 5.3 \\ 
        Point2Roof~\cite{point2roof} & 0.260 & 0.236 & 89.3 & 78.5 & 83.6 & 81.2 & 72.2 & 76.4 & 0.257 & 0.238 & 89.5 & 78.1 & 83.4 & 81.4 & 73.4 & 77.2 \\ 
        PointTransformer$^*$~\cite{point_transformer} & 0.257 & 0.238 & 89.7 & 79.1 & 84.1 & 81.3 & 72.4 & 76.6 & 0.259 & 0.233 & 89.7 & 79.4 & 84.2 & 81.5 & 73.8 & 77.5 \\ 
        PointMLP$^*$~\cite{pointmlp} & 0.255 & 0.233 & \cellcolor{third}{90.6} & 79.0 & 84.4 & 81.6 & 72.8 & 76.9 & 0.255 & 0.229 & 89.8 & 79.7 & 84.4 & 81.4 & 74.1 & 77.6 \\ 
        PointNeXt$^*$~\cite{pointnext} & 0.256 & 0.229 & 90.4 & 79.4 & 84.5 & 82.0 & 73.9 & \cellcolor{third}{78.0} & 0.255 & 0.231 & \cellcolor{third}{91.2} & 79.9 & 85.2  & 82.7 & 74.5 & 78.4 \\ 
        PointMeta$^*$~\cite{pointmeta} & \cellcolor{third}{0.251} & \cellcolor{third}{0.225} & \cellcolor{third}{90.6} & \cellcolor{third}{79.8} & \cellcolor{third}{84.9} & \cellcolor{third}{82.4} & \cellcolor{third}{73.3} & 77.6 & \cellcolor{third}{0.253} & \cellcolor{third}{0.222} & 90.9 & \cellcolor{third}{80.9} & \cellcolor{third}{85.6} & \cellcolor{third}{83.0} & \cellcolor{third}{75.2} & \cellcolor{third}{78.9} \\ 
        BWFormer~\cite{bwformer} & \cellcolor{second}{0.245} & \cellcolor{second}{0.213} & \cellcolor{second}{91.4} & \cellcolor{second}{80.1} & \cellcolor{second}{85.4} & \cellcolor{second}{84.2} & \cellcolor{first}{74.6} & \cellcolor{second}{79.1} & \cellcolor{second}{0.242} & \cellcolor{second}{0.201} & \cellcolor{second}{92.4} & \cellcolor{second}{82.8} & \cellcolor{second}{87.3} & \cellcolor{second}{84.9} & \cellcolor{second}{77.6} & \cellcolor{second}{81.1} \\ 
        \midrule
        Delaunay Canopy & \cellcolor{first}{0.232} & \cellcolor{first}{0.194} & \cellcolor{first}{94.5} & \cellcolor{first}{82.8} & \cellcolor{first}{88.3} & \cellcolor{first}{87.1} & \cellcolor{second}{74.0} & \cellcolor{first}{80.0} & \cellcolor{first}{0.230} & \cellcolor{first}{0.190} & \cellcolor{first}{95.1} & \cellcolor{first}{85.7} & \cellcolor{first}{90.2} & \cellcolor{first}{87.5} & \cellcolor{first}{77.9} & \cellcolor{first}{82.4} \\
        \bottomrule
    \end{tabular}
    }
    \label{tab:main_total}
\end{table*}

\begin{figure*}[t]
    \centering
    \includegraphics[width=0.95\linewidth]
    {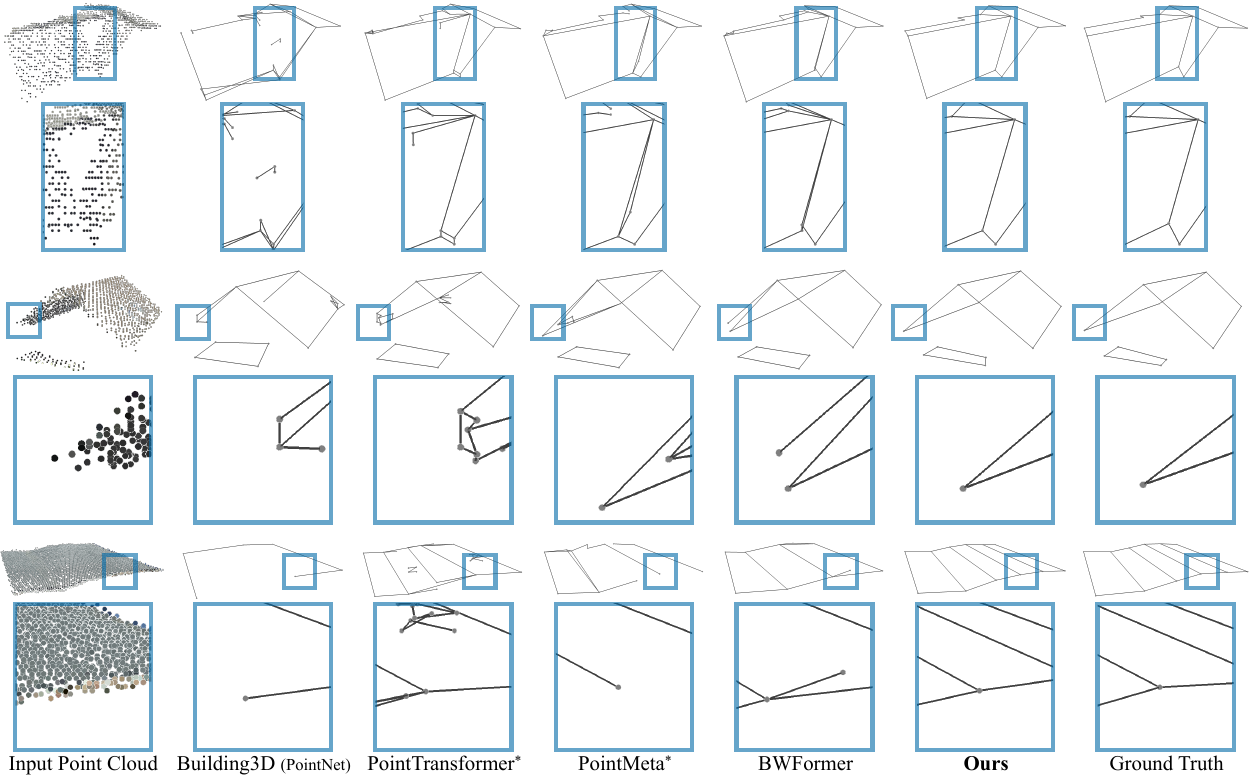}
    \caption{
    \textbf{Qualitative comparison with other methods.}
    The notation $\texttt{MODEL}^*$ signifies a model acting as the feature extractor within the Point2Roof~\cite{point2roof} pipeline.
    The \textcolor{blue}{blue boxes} illustrate the robustness of our framework, demonstrating its capability to reconstruct the wireframe even in regions that are severely sparse (first row), highly noisy (second row), or exhibit minute degrees of curvature (third row).
    }
    \label{fig:qualitative_blue}
\end{figure*}

\begin{figure*}[t]
    \centering
    \includegraphics[width=0.9\linewidth]
    {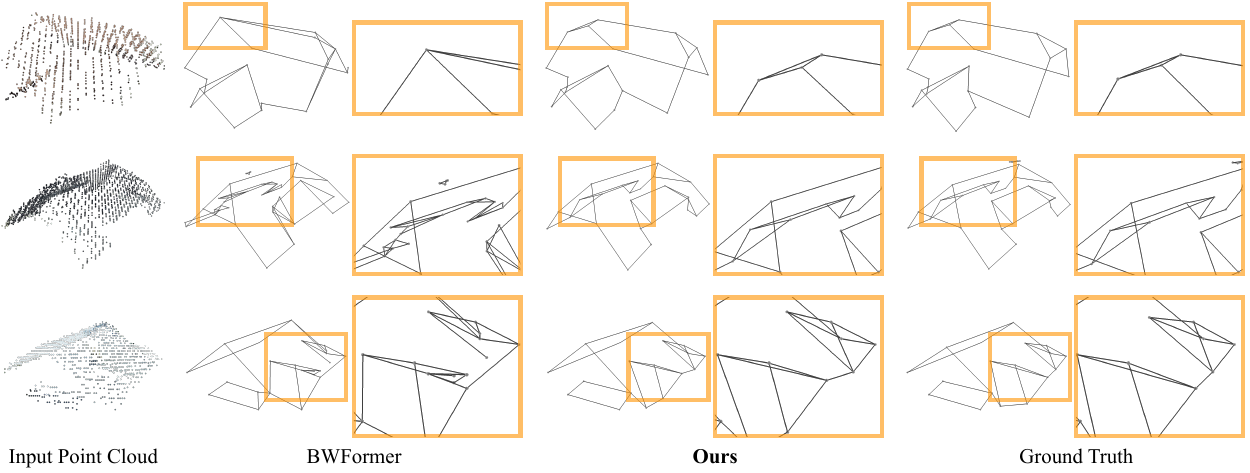}
    \caption{
    \textbf{Qualitative comparison with the strongest baseline BWFormer~\cite{bwformer} on challenging cases with complicated roof structures.}
    The \textcolor{orange}{orange boxes} show our method's ability to fully leverage 3D geometric context, a capability often absent when only 2D information is employed, thus facilitating superior detection of interior corners in addition to those on the periphery.
    }
    \label{fig:qualitative_orange}
\end{figure*}

\subsection{Evaluation Results}

\myparagraph{Quantitative Evaluation.}
We conduct experiments measuring the wireframe reconstruction performance of the Delaunay Canopy against other methods~\cite{point2roof, bwformer} on both the Tallinn city and entry-level datasets from Building3D~\cite{building3d}.
As evident in \Cref{tab:main_total}, our framework performs favorably against other methods on the Tallinn city dataset. \Cref{tab:main_total} further confirms that our framework exhibits superior performance across all metrics on the entry-level dataset. This indicates that our pipeline, by possessing an adaptive search space and precluding unnecessary candidates, possesses an enhanced capacity for selecting valid corners and wires.
Further experimental results, including $K$-fold cross-validation, are provided in the supplementary material.

\myparagraph{Qualitative Analysis.}
\cref{fig:qualitative_blue} shows the qualitative comparison between our framework and recent state-of-the-art methods~\cite{point2roof, bwformer}. 
It is particularly noteworthy that the most challenging aspects of reconstruction—namely, areas that are severely sparse, highly noisy, or where subtle curvature is imperceptible—are notoriously difficult to render faithfully.
Our framework, however, successfully reconstructs these complex, high-difficulty sections.
Furthermore, \cref{fig:qualitative_orange} presents a comparison with BWFormer~\cite{bwformer} on several challenging cases featuring complicated roof structures.
While BWFormer utilizes a height map as input and consequently forfeits 3D geometric context, making it vulnerable in detecting internal corners (those not on the boundary), our framework selects these interior corners with superior efficacy by fully leveraging the available 3D information.
Finally, \cref{fig:non_ideal} illustrates that our approach remains robust even in extreme cases of significant noise and structural incompleteness that go beyond typical challenging samples.
Extended qualitative comparisons and a detailed robustness analysis under diverse challenging scenarios are available in the supplementary.

\setlength{\tabcolsep}{1mm}
\begin{table}[t]
    \begin{minipage}[t]{0.48\linewidth}
        \vspace{0pt}
        \centering
        \includegraphics[width=\linewidth]{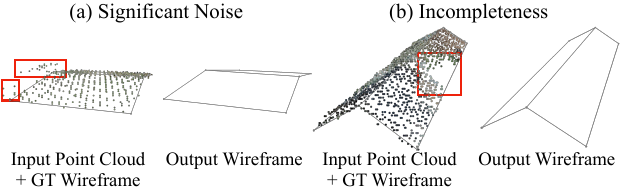}
        \captionof{figure}{\textbf{Robustness to real-world artifacts.} Our method effectively recovers structures even in regions with severe noise or scan voids.}
        \label{fig:non_ideal}
    \end{minipage}
    \hfill
    \begin{minipage}[t]{0.48\textwidth}
        \vspace{0pt}
        \caption{\textbf{Ablation study} on corner score sampling and prior-based query scaling. \colorbox{first}{Best} and \colorbox{second}{second best} are highlighted.}
        \centering
        \setlength{\aboverulesep}{0pt}
        \setlength{\belowrulesep}{0pt}
        \setlength{\tabcolsep}{2pt}
        \resizebox{\columnwidth}{!}{
        \begin{tabular}{cc cc ccc ccc}
            \toprule
            Score & Query & \multicolumn{2}{c}{Distance} & \multicolumn{3}{c}{Corner} & \multicolumn{3}{c}{Wire} \\
            \cmidrule(lr){3-4} \cmidrule(lr){5-7} \cmidrule(lr){8-10}
            Sampling & Scaling & WED $\downarrow$ & ACO $\downarrow$ & CP $\uparrow$ & CR $\uparrow$ & CF1 $\uparrow$ & EP $\uparrow$ & ER $\uparrow$ & EF1 $\uparrow$ \\
            \midrule
            - & - & 0.257 & 0.234 & 88.9 & 78.2 & 83.2 & 81.1 & 72.1 & 76.3 \\
            \checkmark & - & \cellcolor{second}{0.239} & \cellcolor{second}{0.210} & \cellcolor{second}{93.1} & \cellcolor{second}{81.1} & \cellcolor{second}{86.7} & \cellcolor{second}{85.6} & \cellcolor{second}{73.7} & \cellcolor{second}{79.2} \\
            - & \checkmark & 0.242 & 0.223 & 92.4 & 80.3 & 85.9& 84.9 & 72.5 & 78.2\\
            \checkmark & \checkmark & \cellcolor{first}{0.232} & \cellcolor{first}{0.194} & \cellcolor{first}{94.5} & \cellcolor{first}{82.8} & \cellcolor{first}{88.3}& \cellcolor{first}{87.1} & \cellcolor{first}{74.0} & \cellcolor{first}{80.0}\\
            \bottomrule
        \end{tabular}
        }
        \label{tab:ablation}
    \end{minipage}
\end{table}

\subsection{Additional Analyses}

\myparagraph{Ablation Study.}
We conduct an ablation study on the Tallinn city dataset to demonstrate the efficacy of our proposed method, which guides the overall pipeline with a geometric prior via Delaunay graph scoring. We measure how performance shifts based on the intervention of two components: the \textit{corner score sampling} performed during the corner selection process and the \textit{prior-based query scaling} within the wire selection module.
In the setting where corner score sampling is not employed, 150 points were selected as input via Farthest Point Sampling (FPS). Conversely, when prior-based query scaling is not utilized, the process followed the conventional flow of a standard Transformer. As shown in \Cref{tab:ablation}, performance conclusively improves as each method is incrementally introduced, with the setting utilizing all of our proposed methods ultimately achieving the superior performance.

\setlength{\tabcolsep}{1mm}
\renewcommand{\arraystretch}{1.1}
\begin{table}[t]
    \begin{minipage}[t]{0.48\linewidth}
    \caption{Evaluation based on the number of sampled points in corner score sampling. \colorbox{first}{Best} and \colorbox{second}{second best} are highlighted.}
    \vspace{-4pt}
    \centering
    \setlength{\aboverulesep}{4pt}
    \setlength{\belowrulesep}{4pt}
    \setlength{\tabcolsep}{2pt}
    \resizebox{\columnwidth}{!}{
    \begin{tabular}{c cc ccc ccc}
        \toprule
        \# of & \multicolumn{2}{c}{Distance} & \multicolumn{3}{c}{Corner} & \multicolumn{3}{c}{Wire} \\
        \cmidrule(lr){2-3} \cmidrule(lr){4-6} \cmidrule(lr){7-9}
        Points & WED $\downarrow$ & ACO $\downarrow$ & CP $\uparrow$ & CR $\uparrow$ & CF1 $\uparrow$ & EP $\uparrow$ & ER $\uparrow$ & EF1 $\uparrow$ \\
        \midrule
        100 & 0.253 & 0.235 & 89.3 & 80.2 & 84.5& 84.1 & 70.4 & 76.7\\
        125 & 0.249 & \cellcolor{second}{0.219} & 91.4 & \cellcolor{first}{82.9} & \cellcolor{second}{87.0}& \cellcolor{second}{86.2} & 73.9 & 79.6\\
        150 & \cellcolor{first}{0.232} & \cellcolor{first}{0.194} & \cellcolor{first}{94.5} & \cellcolor{second}{82.8} & \cellcolor{first}{88.3}& \cellcolor{first}{87.1} & \cellcolor{second}{74.0} & \cellcolor{first}{80.0}\\
        175 & \cellcolor{second}{0.243} & 0.223 & \cellcolor{second}{92.1} & 80.7 & 86.1& 85.9 & \cellcolor{first}{74.5} & \cellcolor{second}{79.8}\\
        200 & 0.260 & 0.237 & 90.6 & 79.2 & 84.5& 84.7 & 72.2 & 77.9\\
        \bottomrule
    \end{tabular}
    }
    \label{tab:corner_score_sampling}
    \end{minipage}
    \hfill
    \begin{minipage}[t]{0.48\textwidth}
        \caption{Performance variation based on the number of initial layers to which prior-based query scaling is applied. 
        \colorbox{first}{Best} and \colorbox{second}{second best} are highlighted.}
    \vspace{-5pt}
    \centering
    \setlength{\aboverulesep}{0pt}
    \setlength{\belowrulesep}{0pt}
    \setlength{\tabcolsep}{2pt}
    \resizebox{\columnwidth}{!}{
    \begin{tabular}{c cc ccc ccc}
        \toprule
        \# of & \multicolumn{2}{c}{Distance} & \multicolumn{3}{c}{Corner} & \multicolumn{3}{c}{Wire} \\
        \cmidrule(lr){2-3} \cmidrule(lr){4-6} \cmidrule(lr){7-9}
        Layers & WED $\downarrow$ & ACO $\downarrow$ & CP $\uparrow$ & CR $\uparrow$ & CF1 $\uparrow$ & EP $\uparrow$ & ER $\uparrow$ & EF1 $\uparrow$ \\
        \midrule
        0 & 0.239 & 0.210 & 93.1 & 81.1 & 86.8 & 85.6 & 73.7 & 79.2\\
        1 & 0.233& 0.205& 93.5& 81.6& 87.1& 85.8& 73.6& 79.2\\
        2 & \cellcolor{first}{0.230}& \cellcolor{second}{0.197}& \cellcolor{second}{94.2}& 82.2& \cellcolor{second}{87.7}& 86.7& \cellcolor{first}{74.2}& \cellcolor{first}{80.0}\\
        3 & \cellcolor{second}{0.232} & \cellcolor{first}{0.194} & \cellcolor{first}{94.5} & \cellcolor{first}{82.8} & \cellcolor{first}{88.3}& \cellcolor{second}{87.1} & \cellcolor{second}{74.0} & \cellcolor{first}{80.0}\\
        4 & 0.237& 0.204& 93.9& \cellcolor{second}{82.3}& \cellcolor{second}{87.7}& \cellcolor{first}{87.3}& 73.5& \cellcolor{second}{79.8}\\
        5 & 0.246& 0.211& 93.2& 81.9& 87.3& 86.5& 73.1& 79.2\\
        6 & 0.255& 0.213& 92.8& 80.8& 86.4& 85.2& 72.4& 78.3\\
        \bottomrule
    \end{tabular}
    }
    \vspace{-5pt}
    \label{tab:prior_based_query_scaling}
    \end{minipage}
\end{table}

\myparagraph{Analysis of Corner Score Sampling.}
\Cref{tab:corner_score_sampling} illustrates the performance variation of the \textit{corner score sampling}, which specifically depends on the value of $K$, representing the number of vertices with a high corner score selected from the input point cloud.
If an insufficient number of points is filtered, the search space becomes unduly constrained, potentially excluding valid corners from the candidate pool. 
Conversely, filtering too many points enlarges the search space, a limitation observed in prior methods, and reduces the benefit of the geometric prior.
The experiment conclusively demonstrates that the setting filtering 150 vertices yields the most favorable overall performance.

\myparagraph{Analysis of Prior-based Query Scaling.}
\Cref{tab:prior_based_query_scaling} delineates the shift in performance contingent upon the number of layers to which the prior-based query scaling is applied within the wire selection module. Performance progressively improves as the number of layers receiving the scaling is incrementally increased from the initial layer. However, applying query scaling to the later layers risks diluting the semantically rich features that the model has independently generated. Consequently, based on the setting that yields the most excellent performance in \Cref{tab:prior_based_query_scaling}, our framework implements query scaling only across the first three layers.

\myparagraph{Effect of Geometric Prior vs. Architecture.}
To verify that the performance gains stem from our geometric prior rather than the Transformer architecture, we conduct a comparative analysis by substituting our wire selection module. 
As shown in \Cref{tab:prior_vs_arch}, replacing our module (Exp.\#2) with that of Point2Roof~\cite{point2roof} (Exp.\#1) results in only marginal performance shifts. 
In contrast, the introduction of the Delaunay-derived prior via prior-based query scaling (Exp.\#3) yields significant improvements. 
This confirms that the primary contribution lies in the geometric guidance of the Delaunay Canopy rather than mere architectural modifications.

\begin{figure}[t]
    \centering
    \includegraphics[width=0.8\linewidth]{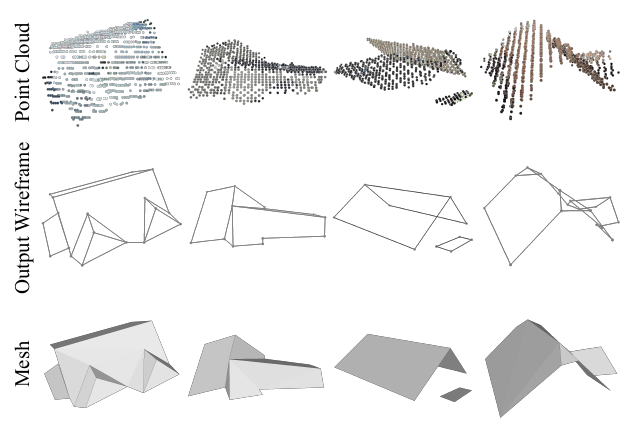}
    \caption{
    The resulting mesh, reconstructed from the wireframe output generated by the Delaunay Canopy.
    The wireframe demonstrably encapsulates sufficient geometric information, allowing the resultant mesh to exhibit a highly faithful form.
    }
    \label{fig:mesh_recon}
\end{figure}

\myparagraph{Extensions and Limitations.}
To further demonstrate the structural fidelity of our reconstructed wireframes, 
we convert them into a mesh format (\cref{fig:mesh_recon}).
While the sparsity and noise of airborne LiDAR data inherently make precise comprehension of building shape challenging, our Delaunay graph scoring method extracts the underlying surface and curvature of the actual building. 
This capability ensures the creation of a geometrically consistent output form.
However, our framework can encounter challenges in cases of data absence, geometric ambiguity, or topologically closed structures. 
A comprehensive analysis and visualization of these failure cases are provided in the supplementary.

\renewcommand{\arraystretch}{1.0}
\begin{table}[t]
    \caption{Analysis of geometric prior vs. architecture.}
    \centering
    \setlength{\aboverulesep}{0pt}
    \setlength{\belowrulesep}{0pt}
    \setlength{\tabcolsep}{2pt}
    \resizebox{\columnwidth}{!}{
    \begin{tabular}{c c c cc ccc ccc}
        \toprule
        Exp. & Wire Selection & Prior-based & \multicolumn{2}{c}{Distance} & \multicolumn{3}{c}{Corner} & \multicolumn{3}{c}{Wire} \\
        \cmidrule(lr){4-5} \cmidrule(lr){6-8} \cmidrule(lr){9-11}
        \# & Module & Query Scaling & WED $\downarrow$ & ACO $\downarrow$ & CP $\uparrow$ & CR $\uparrow$ & F1 $\uparrow$ & EP $\uparrow$ & ER $\uparrow$ & F1 $\uparrow$ \\
        \midrule

        1 & Point2Roof & -
        & 0.248 & 0.225 & 92.4 & 80.5 & 86.0 & 85.8 & 72.8 & 78.8 \\
        
        2 & Ours & -
        & 0.239 & 0.210 & 93.1 & 81.1 & 86.7 & 85.6 & 73.7 & 79.2 \\

        3 & Ours & \checkmark
        & {0.232} & {0.194}
        & {94.5} & {82.8} & {88.3}
        & {87.1} & {74.0} & {80.0} \\
        \bottomrule
    \end{tabular}
    }
    \label{tab:prior_vs_arch}
\end{table}

\section{Conclusion}
\label{sec:conclusion}

In this work, we present Delaunay Canopy, a 3D geometry-driven framework that redefines building wireframe reconstruction from airborne LiDAR point clouds. By introducing the Delaunay Graph as a structural prior, our method transforms the raw, irregular point cloud into a coherent geometric manifold through Delaunay Graph Scoring, yielding rich curvature-aware cues that guide precise corner and wire inference. This principled design not only reduces the ambiguity of sparse 3D search but also overcomes the distortions inherent in 2D projections, enabling the model to reason directly within the true 3D space. Extensive evaluations on the Building3D Tallinn city and Building3D entry-level datasets reveal that Delaunay Canopy achieves striking improvements in structural fidelity and prediction stability, marking a decisive step toward topology-centric, geometry-grounded urban reconstruction.

\bibliographystyle{splncs04}
\bibliography{main}

\clearpage
\setcounter{section}{0} 
\renewcommand{\thesection}{\Alph{section}}
\renewcommand{\thesubsection}{\thesection.\arabic{subsection}}

\section{Further Implementation Specifics}
\label{supp_sec:implementaion_details}

\subsection{Training Details for Corner and Wire Selection}
To train our corner and wire selection modules, we formulate the objective function based on set prediction. 
Let $\mathbf{C}^*$ and $\mathcal{W}^*$ denote the set of ground truth 3D corners and wires, respectively.
Correspondingly, corner selection module and wire selection module predict the set of corners $\tilde{\mathbf{C}}$ and the set of wires $\tilde{\mathcal{W}}$, respectively.
The total loss $\mathcal{L}_\text{total}$ is composed of the corner loss $\mathcal{L}_\text{corner}$ and the wire loss $\mathcal{L}_\text{wire}$.

\subsubsection{Corner Loss.}
Since the corner selection module outputs a set of predictions without intrinsic ordering, we first find the optimal permutation $\hat{\sigma}$ by performing a bipartite matching between the predicted corners $\tilde{\mathbf{C}}$ and the ground truth corners $\mathbf{C}^*$. 
Thus, we minimize the matching cost $\mathcal{L}_\text{match}$ using the Cross-Entropy ($\text{CE}$) and the Manhattan distance ($D_{\text{Manhattan}}$) as
\begin{equation}
    \mathcal{L}_\text{match}(c^*_i, \tilde{c}_{\sigma(i)}) =  \alpha \text{CE}(c^*_i, \tilde{c}_{\sigma(i)}) + \beta D_{\text{Manhattan}}(c^*_i, \tilde{c}_{\sigma(i)}),
\end{equation}
where $c_i^*$ is the $i$-th ground-truth corner, and $\tilde{c}_{\sigma(i)}$ the $i$-th predicted corner under the influence of the bipartite-matching permutation $\sigma$. The weights $\alpha$ and $\beta$ are hyperparameters balancing two terms.
Based on the optimal permutation $\hat{\sigma}$ that minimizes the matching cost, the \textit{Corner Loss} ($\mathcal{L}_\text{corner}$) is computed as
\begin{equation}
    \mathcal{L}_\text{corner} = \frac{1}{|\mathbf{C}^*|}\sum_{i=1}^{|\mathbf{C}^*|}  \alpha \text{CE}(c^*_i, \tilde{c}_{\hat{\sigma}(i)}) + \beta D_{\text{Manhattan}}(c^*_i, \tilde{c}_{\hat{\sigma}(i)}).
\end{equation}

\subsubsection{Wire Loss.}
The wire selection module serves to adjudicate the structural validity of every wire candidate presented. For the predicted set of wires $\tilde{\mathcal{W}}$, we compute the binary cross-entropy loss against the ground truth wires $\mathcal{W}^*$. 
The \textit{Wire Loss} ($\mathcal{L}_\text{wire}$) is defined as:
\begin{equation}
    \mathcal{L}_{wire} = \text{CE}(\mathcal{W}^*, \tilde{\mathcal{W}}).
\end{equation}
This loss term penalizes the model for incorrect wire selections, ensuring that the topological structure of the building is accurately reconstructed.

\subsubsection{Total Loss.}
Finally, the total objective function is a weighted sum of the aforementioned losses:
\begin{equation}
    \mathcal{L}_\text{total} = \mathcal{L}_\text{corner} + \gamma \mathcal{L}_\text{wire},
\end{equation}
where $\gamma$ is balancing coefficients for the corner and wire losses.

\subsection{Details of Delaunay Graph Scoring}

\subsubsection{Delaunay Triangulation.} 
\begin{center}
\begin{minipage}{0.85\linewidth}
\begin{algorithm}[H]
\caption{Delaunay Triangulation via Incremental Insertion}
\label{alg:delaunay_incremental}
\SetKwInOut{KwIn}{Input}
\SetKwInOut{KwOut}{Output}

\KwIn{Set of points $V = \{v_1, \dots, v_N\}$}
\KwOut{Delaunay Triangulation $\mathcal{T}$ (Set of Simplices)}

\textbf{Initialize $\mathcal{T}$}: Create an initial large bounding simplex $S_{\text{init}}$ that contains all points in $V$\;

\For{$i = 1$ \KwTo $N$}{
    Let $v_i$ be the point to insert\;
    \If{$v_i$ is already a vertex of $\mathcal{T}$}{
        \textbf{continue}\;
    }
    
    \textbf{Find Conflict Region}: Identify the set of simplices $\mathcal{C} \subset \mathcal{T}$ whose circumcircles contain $v_i$\;
    
    \If{$\mathcal{C} = \emptyset$}{
        \textbf{continue}\;
    }
    
    \textbf{Compute Boundary}: Determine the boundary $\partial \mathcal{C}$ of the conflict region $\mathcal{C}$\;
    
    \textbf{Remove Conflict Simplices}: $\mathcal{T} \leftarrow \mathcal{T} \setminus \mathcal{C}$\;
    
    \textbf{Create New Simplices}: For each edge $\mathbf{e} \in \partial \mathcal{C}$ on the boundary, create a new simplex $S_{\text{new}}$ using $\mathbf{e}$ and the new point $v_i$\;
    
    $\mathcal{T} \leftarrow \mathcal{T} \cup \{S_{\text{new}} \mid \mathbf{e} \in \partial \mathcal{C}\}$\;
    
    \textbf{Optional Optimization (Flip Test)}: For each new simplex $S_{\text{new}}$ and its neighbors, perform a local check to ensure the Delaunay condition is maintained (e.g., edge flipping)\;
}

\textbf{Remove Bounding Simplex}: Delete all simplices connected to the initial bounding simplex $S_{\text{init}}$ vertices from $\mathcal{T}$\;

\Return $\mathcal{T}$\;
\end{algorithm}
\end{minipage}
\end{center}
\textit{Delaunay graph scoring} is a method that is predicated on receiving an airborne LiDAR point cloud of a scanned structure. 
This process begins with the requisite application of \textit{Delaunay Triangulation} to this input point cloud.
The Delaunay Triangulation is a fundamental tessellation that maximizes the minimum angle of all triangles in the partitioning, thereby ensuring mesh quality. 
The core principle adheres to the empty circumsphere criterion: for any simplex in the triangulation, the circumsphere defined by its vertices must contain no other points from the point cloud.
In this work, the Delaunay Triangulation is computed using the \texttt{scipy.spatial.Delaunay} implementation, which relies on the established \texttt{Qhull} library. 
The detailed procedure can be found in \cref{alg:delaunay_incremental}.
Following this algorithm, the resulting set of simplices $\mathcal{T}$ is then approached from a graph-theoretic perspective, allowing us to extract the geometric information for each constituent element from the resultant Delaunay graph $G = (V, E, F)$.

\subsubsection{Wire-Wise Path Score.}
To guide the discrimination of wire candidates based on their likelihood of validity within the wire selection module, we compute the \textit{wire-wise path score}, a metric reflective of the candidate's structural validity.
In a more detailed fashion, for a given wire candidate $w_{ij}$, we first determine the shortest path, $P_{i \rightarrow j}^G$, which interconnects its two endpoints, $v_i$ and $v_j$, within the Delaunay graph $G = (V, E, F)$. The final metric is then derived by aggregating the dihedral angles of all edges that lie along this designated path. Crucially, this shortest path is elucidated using Dijkstra's algorithm~\cite{dijkstra} on the Delaunay graph $G = (V, E, F)$, where the path cost is parameterized by the actual Euclidean distance between the constituent points.
\section{Additional Evaluation Results}
\label{supp_sec:additional_evaluation_results}

To further substantiate the efficacy of \textit{Delaunay Canopy} for building wireframe reconstruction, we provide supplementary evaluation experiments and accompanying analysis. 
The comparative results on the official Building3D leaderboard are presented in \cref{suppsubsubsec:official_leaderboard}, and the quantitative results derived from $K$-fold cross-validation are fully detailed within \cref{suppsubsubsec:additional_quantitative_results}. 
Furthermore, an expanded collection of wireframe outputs and their corresponding qualitative analyses are presented in \cref{suppsubsubsec:additional_qualitative_results}.

\subsection{Performance on the Official Building3D Leaderboard}
\label{suppsubsubsec:official_leaderboard}
\begin{table}[h]
    \caption{
    Performance evaluation based on the official Building3D~\cite{building3d} Tallinn city dataset leaderboard. Baseline metrics are sourced from the public leaderboard. The notation $\texttt{MODEL}^\dagger$ designates models acting exclusively as the feature extraction backbone within the standard Building3D~\cite{building3d} architecture. 
    \colorbox{first}{Best}, \colorbox{second}{second best}, and \colorbox{third}{third best} are highlighted.
    }
    \centering
    \setlength{\aboverulesep}{2pt}
    \setlength{\belowrulesep}{2pt}
    \setlength{\tabcolsep}{8pt}
    \resizebox{\columnwidth}{!}{
    \begin{tabular}{l cc ccc ccc}
        \toprule
        \multirow{2}{*}{Method} & \multicolumn{2}{c}{Distance} & \multicolumn{3}{c}{Corner} & \multicolumn{3}{c}{Wire} \\
        \cmidrule(lr){2-3} \cmidrule(lr){4-6} \cmidrule(lr){7-9}
        & WED $\downarrow$ & ACO $\downarrow$ & CP $\uparrow$ & CR $\uparrow$ & CF1 $\uparrow$ & EP $\uparrow$ & ER $\uparrow$ & EF1 $\uparrow$ \\
        \midrule
        PointMAE$^\dagger$~\cite{pointmae} & - & 0.330 & 75.0 & 47.0 & 58.0 & 52.0 & 12.0 & 20.0 \\
        PointM2AE$^\dagger$~\cite{pointm2ae} & - & 0.320 & 79.0 & 58.0 & 67.0 & 50.0 & 7.0 & 12.0 \\
        Point2Roof~\cite{point2roof} & - & 0.390 & 65.0 & 30.0 & 41.0 & 66.0 & 8.0 & 14.0 \\
        Linear self-supervised~\cite{building3d} & - & 0.350 & 70.0 & 60.0 & 65.0 & 67.0 & 16.0 & 25.0 \\
        Supervised~\cite{building3d} & - & 0.290 & 90.0 & 53.0 & 66.0 & \cellcolor{second}{88.0} & 23.0 & 36.0 \\
        PC2WF~\cite{pc2wf} & - & 0.520 & 18.0 & 67.0 & 28.0 & 2.0 & 15.0 & 1.0 \\
        PBWR~\cite{pbwr} & \cellcolor{third}{0.271} & \cellcolor{third}{0.222} & \cellcolor{first}{98.5} & \cellcolor{third}{68.8}& \cellcolor{third}{81.0} & \cellcolor{first}{94.3} & \cellcolor{third}{65.4} & \cellcolor{third}{77.2} \\
        BWFormer~\cite{bwformer} & \cellcolor{second}{0.238} & \cellcolor{first}{0.204} & \cellcolor{third}{94.9} & \cellcolor{second}{82.7}& \cellcolor{second}{88.4} & 85.5 & \cellcolor{second}{74.1} & \cellcolor{second}{79.4} \\
        \midrule
        Delaunay Canopy & \cellcolor{first}{0.225} & \cellcolor{second}{0.206} & \cellcolor{second}{95.7} & \cellcolor{first}{83.9} & \cellcolor{first}{89.4} & \cellcolor{third}{87.6} & \cellcolor{first}{75.4} & \cellcolor{first}{81.1} \\
        \bottomrule
    \end{tabular}
    }
    \label{tab:official_leaderboard}
\end{table}
To ensure a rigorous and standardized comparison, we evaluate our framework according to the official Building3D~\cite{building3d} leaderboard protocol. 
As shown in \cref{tab:official_leaderboard}, we compare \textit{Delaunay Canopy} with various state-of-the-art methods and established baselines based on the leaderboard. Our framework not only surpasses the methods utilized as feature extractors (denoted with $^\dagger$) but also demonstrates superior structural fidelity compared to recent leading approaches such as PBWR~\cite{pbwr} and BWFormer~\cite{bwformer}. This confirms that the geometric prior embedded within the Delaunay Canopy provides a generalized, state-of-the-art capability in wireframe reconstruction under official benchmark standards.

\subsection{Additional Quantitative Results ($K$-Fold Cross-Validation)}
\label{suppsubsubsec:additional_quantitative_results}
\begin{table}[h]
    \caption{Robustness validation via 5-Fold Cross-Validation. The notation $\texttt{MODEL}^*$ signifies a model acting as the feature extractor within the Point2Roof~\cite{point2roof} pipeline. \colorbox{first}{Best} and \colorbox{second}{second best} are highlighted.}
    \centering
    \setlength{\aboverulesep}{2pt}
    \setlength{\belowrulesep}{2pt}
    \setlength{\tabcolsep}{10pt}
    \resizebox{\columnwidth}{!}{
    \begin{tabular}{l cc ccc ccc}
        \toprule
        \multirow{2}{*}{Method} & \multicolumn{2}{c}{Distance} & \multicolumn{3}{c}{Corner} & \multicolumn{3}{c}{Edge} \\
        \cmidrule(lr){2-3} \cmidrule(lr){4-6} \cmidrule(lr){7-9}
        & WED $\downarrow$ & ACO $\downarrow$ & CP $\uparrow$ & CR $\uparrow$ & CF1 $\uparrow$ & EP $\uparrow$ & ER $\uparrow$ & EF1 $\uparrow$ \\
        \midrule
        Building3D (PointNet)~\cite{pointnet}& 0.271& 0.245& 87.8& 77.9& 82.5& 80.8& 71.5& 75.9\\
        PC2WF~\cite{pc2wf}& 0.548& 0.512& 21.9& 50.1& 30.4& 3.1& 16.5& 5.3\\
        Point2Roof~\cite{point2roof}& 0.265& 0.239& 89.0& 78.2& 83.3& 81.5& 71.8& 76.3\\
        PointTransformer$^*$~\cite{point_transformer}& 0.260& 0.242& 89.4& 78.8& 83.8& 81.7& 72.0& 76.5\\
        PointMLP$^*$~\cite{pointmlp}& 0.258& 0.236& 90.3& 79.3& 84.4& 82.1& 72.5& 77.0\\
        PointNeXt$^*$~\cite{pointnext}& 0.259& 0.232& 90.7& 79.7& 84.8& 82.5& 73.0& 77.5\\
        PointMeta$^*$~\cite{pointmeta}& 0.254& 0.228& 91.0& 80.2& 85.3& 82.9& 73.5& 77.9\\
        BWFormer~\cite{bwformer}& \cellcolor{first}{0.248}& \cellcolor{second}{0.217} & \cellcolor{second}{91.8} & \cellcolor{first}{80.5}& \cellcolor{second}{85.8}& \cellcolor{second}{84.5}& \cellcolor{second}{74.3}& \cellcolor{second}{79.1} \\
        \midrule
        Delaunay Canopy& \cellcolor{second}{0.250}& \cellcolor{first}{0.199}& \cellcolor{first}{93.8}& \cellcolor{second}{79.9} & \cellcolor{first}{86.2}& \cellcolor{first}{86.8}& \cellcolor{first}{74.5}& \cellcolor{first}{80.2}\\
        \bottomrule
    \end{tabular}
    }
    \label{tab:k_fold}
\end{table}
While our initial evaluation utilizes a conventional random train-test split, this approach may risks introducing statistical bias and may fail to fully capture the performance generalization capacity of the model. To ensure the robustness and statistical reliability of our results, we adopt 5-fold cross-validation on the combined training and test sets. This procedure systematically partitions the dataset into five equal subsets, where each subset serves as the validation set exactly once, mitigating the sensitivity of the final performance metric to any single, arbitrary data split. 
\cref{tab:k_fold} attests to the fact that the Delaunay Canopy exhibits a distinct and favorable performance advantage over other competing methodologies.

\subsection{Additional Qualitative Results}
\label{suppsubsubsec:additional_qualitative_results}
\begin{figure}[h]
    \centering
    \includegraphics[width=\linewidth]{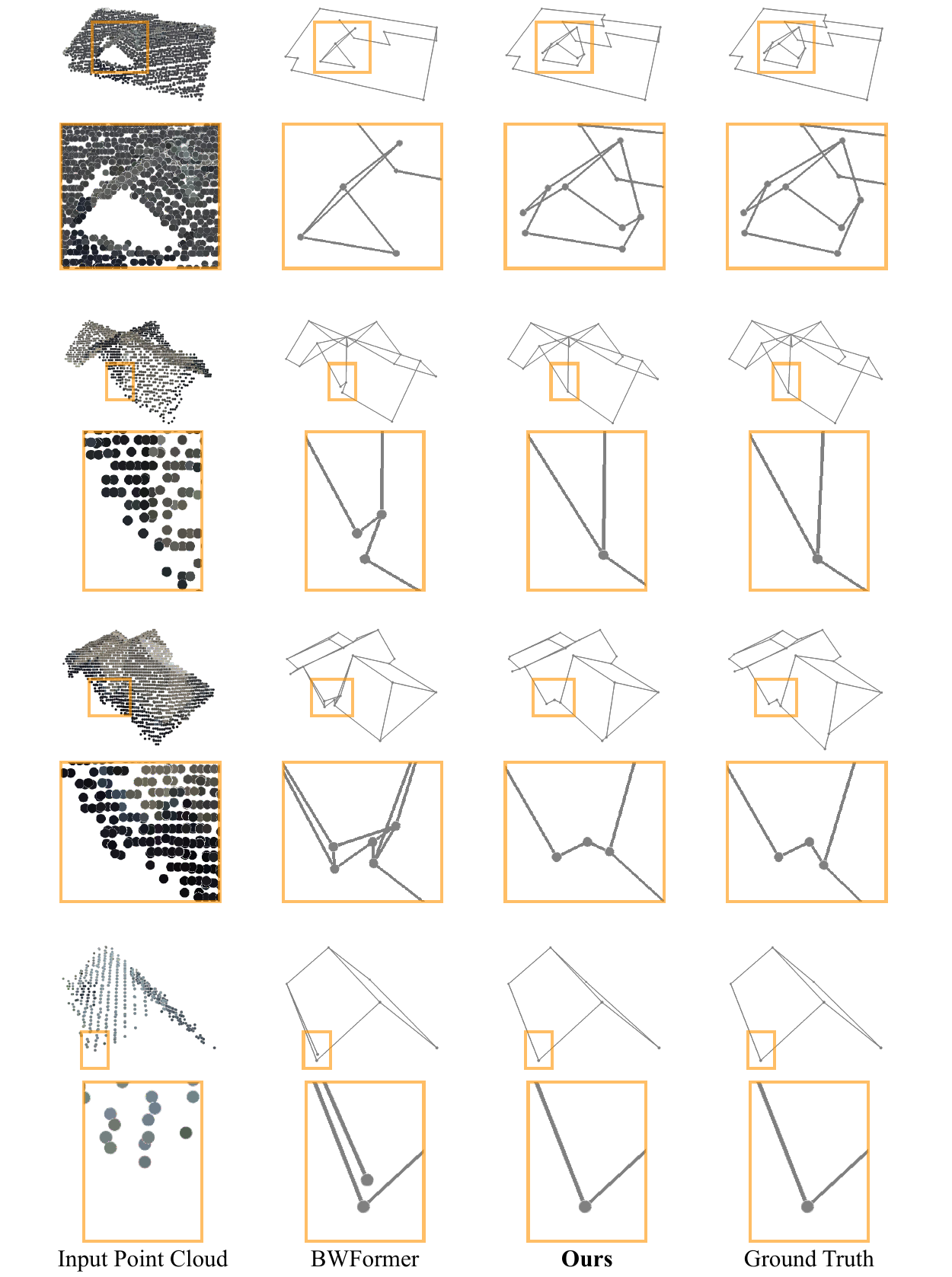}
    \caption{
    Qualitative comparison with the strongest baseline BWFormer~\cite{bwformer}. 
    The orange bounding boxes specifically denote the regions where our method achieves a demonstrably superior reconstruction result.
    }
    \label{fig:additional_visual_results1}
\end{figure}

\begin{figure}[h]
    \centering
    \includegraphics[width=\linewidth]{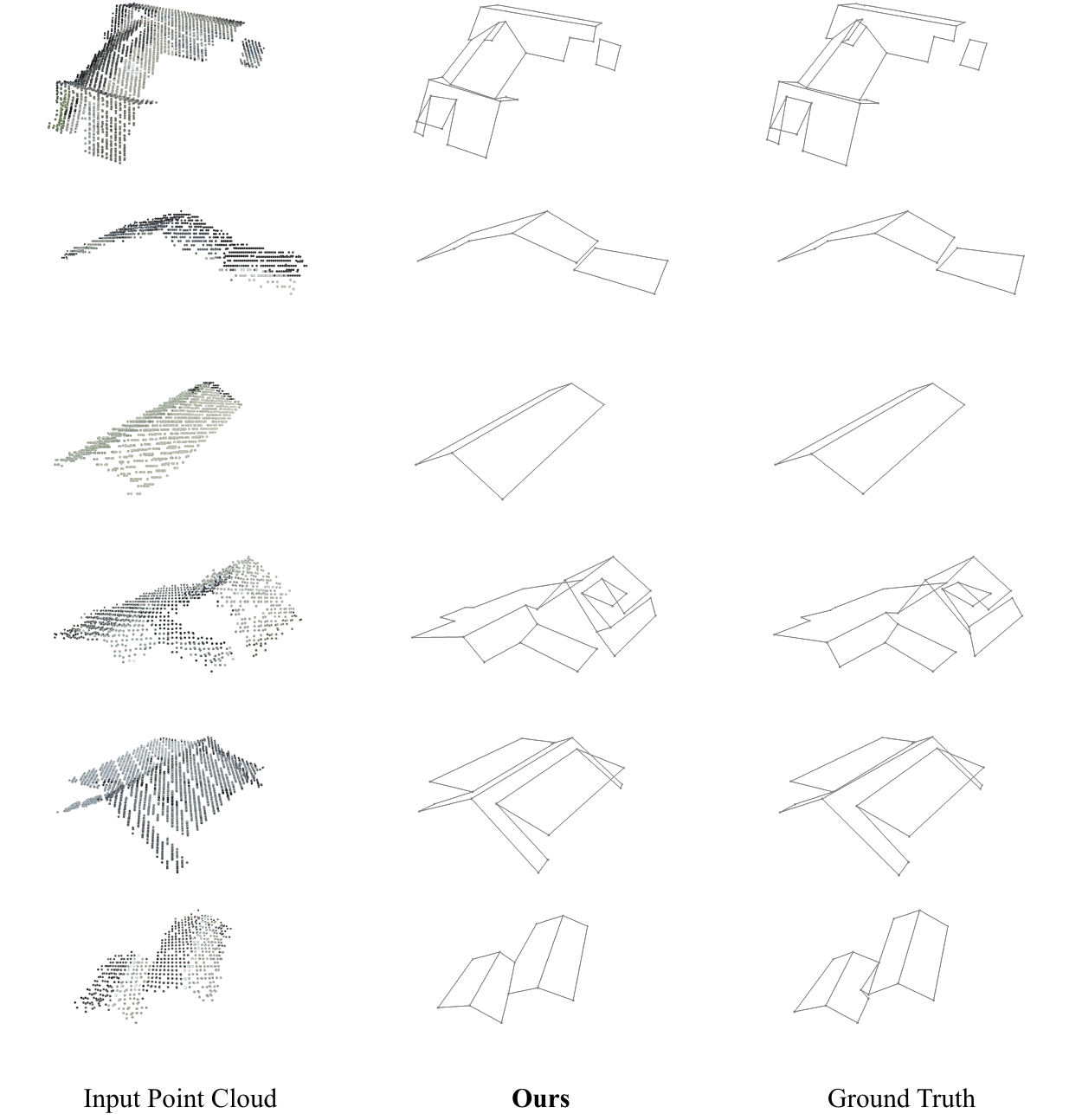}
    \caption{
    Wireframe reconstruction results of Delaunay Canopy on Building3D dataset.
    }
    \label{fig:additional_visual_results2}
\end{figure}

\begin{figure}[h]
    \centering
    \includegraphics[width=\linewidth]{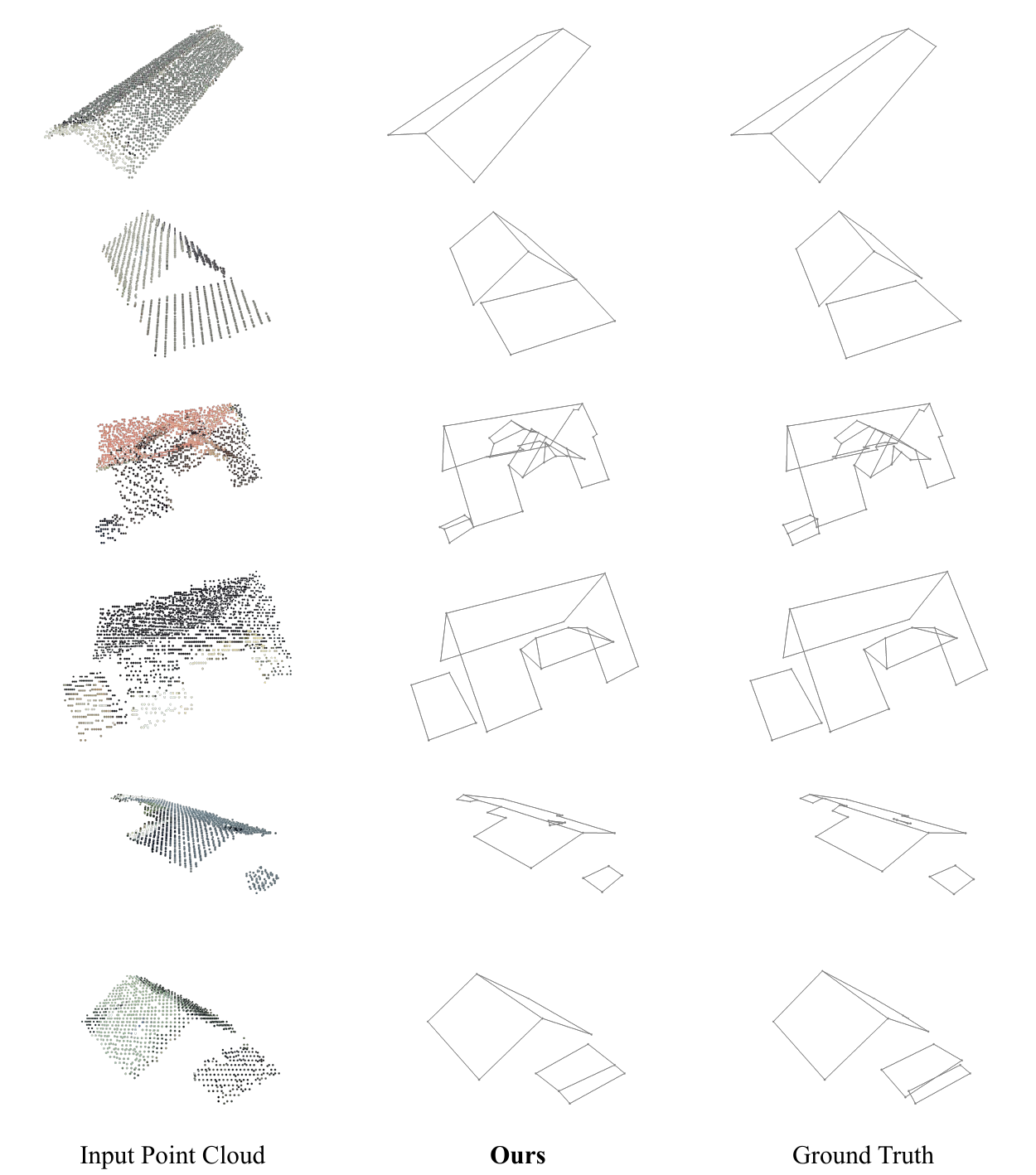}
    \caption{
    Wireframe reconstruction results of Delaunay Canopy on Building3D dataset.
    }
    \label{fig:additional_visual_results3}
\end{figure}

We provide supplementary visualizations of the outcomes wherein airborne LiDAR building point clouds are transformed into a wireframe via \textit{Delaunay Canopy}. \cref{fig:additional_visual_results1,fig:additional_visual_results2,fig:additional_visual_results3} attests to the framework's capacity to competently and robustly reconstruct the wireframe across a spectrum of challenging scenarios, including those marked by acute sparsity, pervasive noise, and point clouds exhibiting subtle curvature. Moreover, a clear advantage is demonstrated by Delaunay Canopy in its capacity to accurately reconstruct the internal corner features, an undertaking that the strongest baseline, BWFormer~\cite{bwformer}, is unable to accomplish with a commensurate level of precision.

\section{Additional Analyses}
\label{supp_sec:additional_analyses}

\subsection{Computational Speed of Delaunay Graph Scoring}
To proactively mitigate potential concerns regarding the computational overhead of our core approach, \textit{Delaunay graph scoring}, the empirical processing time on the Building3D dataset~\cite{building3d} is rigorously measured and presented.
When evaluated on the Building3D Tallinn city train dataset, the mean processing time per point cloud is a mere \textbf{0.367 seconds}. This throughput translates to the ability to execute the Delaunay graph scoring on approximately \textbf{2.7 point clouds per second}.
This performance firmly establishes the approach as a highly efficient and computationally meritorious strategy, especially when factoring in that the 3D building point clouds in the dataset typically encompass several thousand, often exceeding ten thousand, constituent points.

\subsection{Robustness in Challenging Scenarios.}
\begin{table}[h]
    \caption{Robustness validation assessed against a perturbed dataset characterized by augmented sparsity and noise. The notation $\texttt{MODEL}^*$ signifies a model acting as the feature extractor within the Point2Roof~\cite{point2roof} pipeline.} 
    \centering
    \setlength{\aboverulesep}{0pt}
    \setlength{\belowrulesep}{0pt}
    \setlength{\tabcolsep}{10pt}
    \resizebox{\columnwidth}{!}{
    \begin{tabular}{l cc ccc ccc}
        \toprule
        \multirow{2}{*}{Method} & \multicolumn{2}{c}{Distance} & \multicolumn{3}{c}{Corner} & \multicolumn{3}{c}{Edge} \\
        \cmidrule(lr){2-3} \cmidrule(lr){4-6} \cmidrule(lr){7-9}
        & WED $\downarrow$ & ACO $\downarrow$ & CP $\uparrow$ & CR $\uparrow$ & CF1 $\uparrow$ & EP $\uparrow$ & ER $\uparrow$ & EF1 $\uparrow$ \\
        \midrule
        Building3D (PointNet)~\cite{pointnet}& 0.407& 0.354& 78.1& 68.9& 73.2& 72.6& 64.2& 68.2\\
        PC2WF~\cite{pc2wf}& 0.658& 0.612& 14.2& 38.7& 20.8& 1.6& 10.4& 2.7\\
        Point2Roof~\cite{point2roof}& 0.384& 0.341& 79.5& 70.3& 74.6& 73.9& 65.4& 69.4\\
        PointTransformer$^*$~\cite{point_transformer}& 0.367& 0.325& 80.8& 71.5& 75.9& 75.2& 66.5& 70.6\\
        PointMLP$^*$~\cite{pointmlp}& 0.359& 0.318& 81.3& 72.0& 76.4& 75.6& 67.1& 71.1\\
        PointNeXt$^*$~\cite{pointnext}& 0.355& 0.311& 82.1& 72.8& 77.2& 76.3& 68.6& 72.3\\
        PointMeta$^*$~\cite{pointmeta}& 0.348& 0.302& 83.5& 73.6& 78.3& 77.1& 69.3& 73.1\\
        BWFormer~\cite{bwformer}& 0.339& 0.291& 84.4& 74.5& 79.1& 78.7& 70.1& 74.2\\
        \midrule
        Delaunay Canopy& 0.301& 0.263& 88.2& 78.1& 82.8& 80.6& 71.3& 75.7\\
        \bottomrule
    \end{tabular}
    }
    \label{tab:challenging_scenarios}
\end{table}

\begin{figure}[h]
    \centering
    \includegraphics[width=\linewidth]{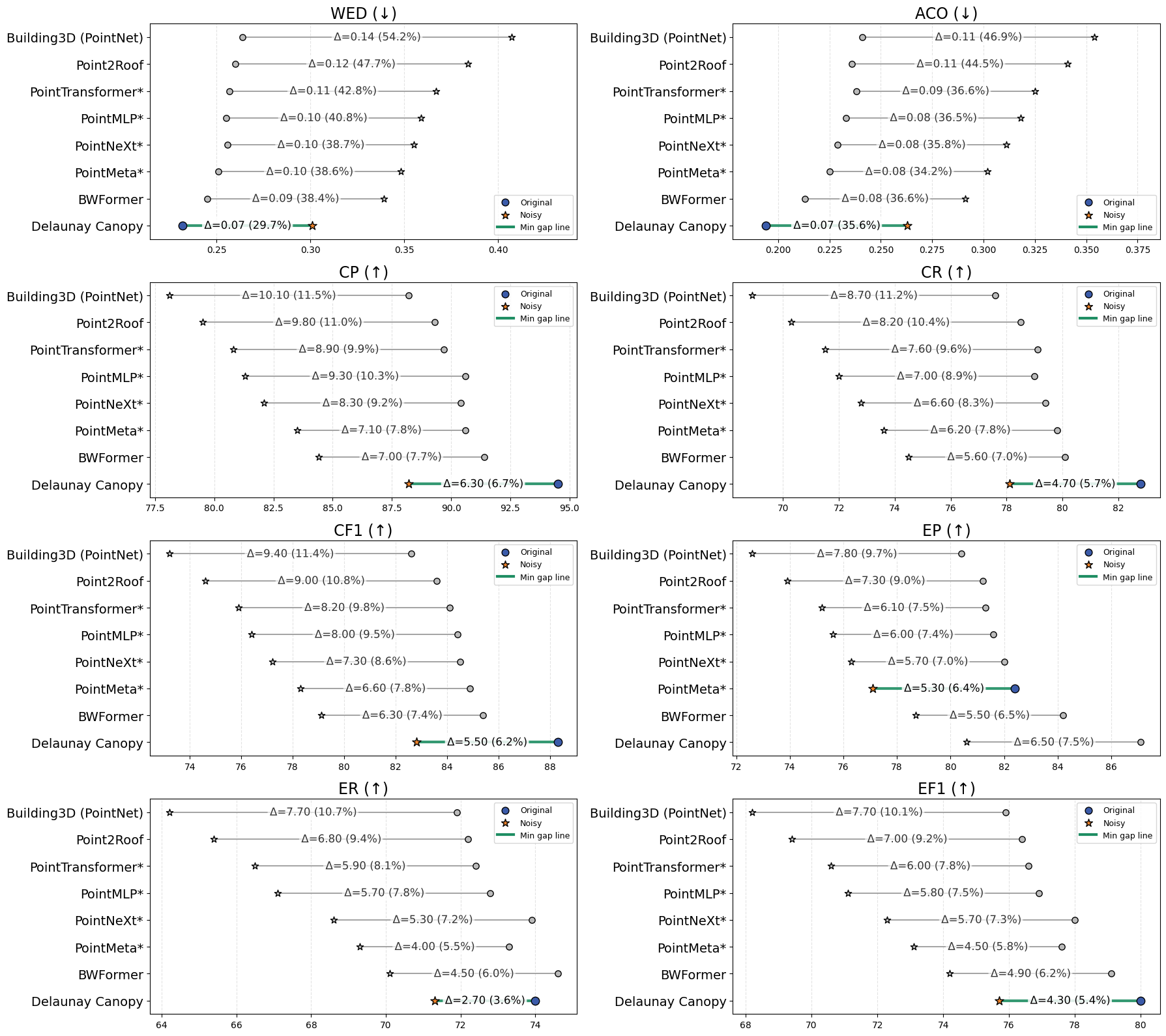}
    \caption{
    Visualization of performance differences between challenging (perturbed point cloud) and original inputs.
    The number next to $\Delta$ indicates the absolute performance difference, and the percentage (\%) shows the relative decrease compared to the original performance.
    The notation $\texttt{MODEL}^*$ signifies a model acting as the feature extractor within the Point2Roof~\cite{point2roof} pipeline.
    }
    \label{fig:robust}
\end{figure}
We compare the robustness of our framework, \textit{Delaunay Canopy}, against the baselines~\cite{pointnet,pc2wf,point2roof,point_transformer,pointmlp,pointnext,pointmeta,bwformer} introduced in the main paper by creating a more challenging test set than the existing Building3D~\cite{building3d} Tallinn city and entry-level datasets. 
For this evaluation, we utilize a custom test set split from the Tallinn city dataset, as mentioned in the main paper, but drastically increase its sparsity and noise compared to the original. 
To induce severe localized sparsity across all point clouds, we employ a controlled removal strategy: in each point cloud, a single point was randomly selected, and this point, along with its nearest neighbors constituting 5\% of the total points in that cloud, was subsequently removed.
Furthermore, to ensure a globally noisy state in the input point clouds, we add Gaussian noise to every point cloud within the test set.

As evinced by \cref{fig:robust} and \Cref{tab:challenging_scenarios}, the Delaunay Canopy exhibits a pronounced advantage over other baselines concerning robustness. This superior performance is attributable to its methodology: by gleaning the intrinsic information of the point cloud's underlying surface through the Delaunay graph, it facilitates geometric extrapolation even into zones lacking explicit data points. This very mechanism, which allows the framework to discern the overarching structural trend from local curvature despite the presence of stochastic perturbations, is what ultimately enables a resilient and robust wireframe reconstruction.
\section{Discussion}
\label{supp_sec:discussion}

\subsection{Limitations}

\begin{figure}[h]
    \centering
    \includegraphics[width=\linewidth]{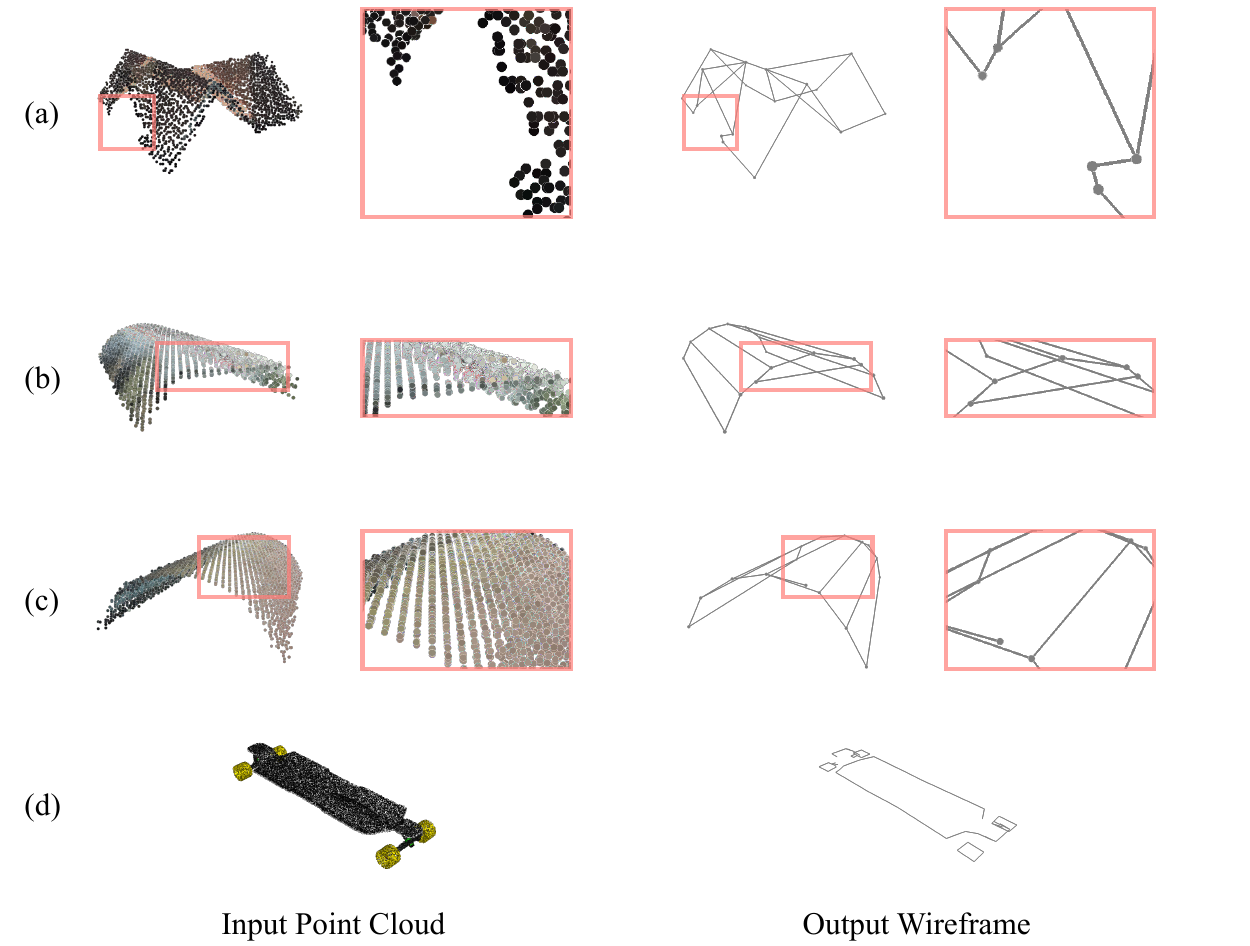}
    \caption{
    Visualization of failure cases. 
    Precise wireframe reconstruction is demonstrably compromised under several key scenarios. (a) When critical regions containing valid corner and wire features are unscanned and consequently absent from the point cloud. (b \& c) When the intrinsic surface topology of the point cloud exhibits excessive smoothness. (d) When processing objects characterized by a closed topology, such as non-building volumetric structures.
    }
    \label{fig:limitations}
\end{figure}

While \textit{Delaunay Canopy} exhibits robust performance across diverse scenarios, it is not without its limitations, presenting opportunities for refinement. 
The overall pipeline is comprised of the corner selection stage and the wire selection stage, each employing a dedicated mechanism to ``\textit{select}'' valid corners and wires from their respective candidates. 
Consequently, a critical challenge emerges when the input LiDAR point cloud fails to capture the requisite valid corner and wire features, \ie, when crucial critical points are omitted during scanning and the corresponding geometric region is severely sparse. 
Under such conditions, the pipeline's capacity to effectively detect both corners and wires is significantly diminished.
A direct inspection of \cref{fig:limitations}(a) intuitively demonstrates that in instances where regions where a valid corner is structurally mandated remain unscanned and are consequently absent from the point cloud, the corner selection mechanism fails to correctly identify the requisite feature during the wireframe reconstruction pipeline. This failure, in turn, results in the resultant wireframe exhibiting suboptimal quality.

A second limitation arises when the intrinsic surface of the point cloud is devoid of sharp angular features, rendering the identification of true wireframe corners and wires inherently ambiguous. In such circumstances, the Delaunay Canopy framework is similarly hindered in its capacity for precise wireframe reconstruction.
\cref{fig:limitations}(b) and \cref{fig:limitations}(c) visually confirms that in regions characterized by rounded surfaces, the framework experiences erroneous corner selection and wire selection.

The final and most principal methodological constraint is the fundamental assumption that Delaunay graph scoring requires input point clouds to be limited to topologically open structures (\eg, non-closed surfaces like building rooftops).
This is because the method strictly requires that only a single $z$-coordinate exists for any given ground coordinate $(x, y)$. Consequently, our methodology cannot be directly or effectively transposed to point clouds captured from topologically closed (volumetric) objects (\cref{fig:limitations}(d)).

\subsection{Future Works}
\textit{Delaunay Canopy} constitutes a framework engineered for the reconstruction of wireframes from 3D building point clouds. Its efficacy stems from utilizing the core approach, \textit{Delaunay graph scoring}, which first yields a robust approximation of the intrinsic surface of the input cloud, then subsequently extracts local curvature information. This information is utilized as a geometric prior to provide critical guidance to the reconstruction pipeline. Crucially, this approach is not solely confined to the architectural domain; it is highly extensible to diverse subjects encountered in autonomous driving and interior scanning applications. Consequently, our primary avenue for future investigation is the generalization of this approach across these disparate fields, culminating in the development of a truly universal wireframe reconstruction framework.

\clearpage
\clearpage

\end{document}